\documentclass{article}

\usepackage{preprint_style,times}

\usepackage{amsmath,amsfonts,bm}

\def\eqref#1{equation~\ref{#1}}

\def\1{\bm{1}}

\def\vx{{\bm{x}}}
\def\vy{{\bm{y}}}

\def\mBeta{{\bm{\beta}}}

\DeclareMathAlphabet{\mathsfit}{\encodingdefault}{\sfdefault}{m}{sl}
\SetMathAlphabet{\mathsfit}{bold}{\encodingdefault}{\sfdefault}{bx}{n}

\usepackage{amsmath}
\usepackage{amsthm}
\usepackage{booktabs}
\usepackage{graphicx}
\usepackage{float}
\usepackage{placeins}
\usepackage{wrapfig}
\usepackage[most]{tcolorbox}
\usepackage{microtype}
\usepackage{url}
\usepackage{hyperref}

\hypersetup{
  colorlinks=true,
  linkcolor=blue,
  citecolor=blue,
  urlcolor=blue
}

\renewcommand{\topfraction}{0.68}
\renewcommand{\bottomfraction}{0.25}
\renewcommand{\textfraction}{0.12}
\renewcommand{\floatpagefraction}{0.72}
\title{Emergent Inverse-Depth Scaling From Nonlinearity In Attention}
\author{%
Zirui Peng\textsuperscript{1}\thanks{Work done during an internship at MIT.},\enspace
Yizhou Liu\textsuperscript{2},\enspace
Ziming Liu\textsuperscript{3},\enspace
Jeff Gore\textsuperscript{2,\textdagger}\\[2pt]
\normalfont\small \textsuperscript{1}Department of Physics, Peking University\\
\normalfont\small \textsuperscript{2}Department of Physics, Massachusetts Institute of Technology, Cambridge, MA\\
\normalfont\small \textsuperscript{3}College of AI, Tsinghua University\\
\normalfont\small \textsuperscript{\textdagger}Corresponding author: \texttt{gore@mit.edu}
}

\begin{document}

\maketitle
\lhead{Preprint}

\begin{abstract}
Scaling laws describe power-law improvements in model performance with
dataset size and parameter count, yet their underlying mechanisms are not
fully understood. To explain the parameter count scaling, existing
theory posits power-law scaling with model depth. In linear-attention models,\footnote{Here, linear attention means that attention scores are computed directly as query--key inner products and that the model contains no nonlinear components.}
this scaling is tied to a power-law data spectrum: unable to selectively
attend to relevant tokens, these models learn according to global spectral
strength, with stronger directions learned before weaker ones. Large language
models, however, can be strongly nonlinear. Here, we show that nonlinear
attention yields inverse-depth decay of loss across all tested data
spectra. Nonlinearity enables attention to focus selectively on relevant
tokens, allowing strong and weak spectral directions to be learned in
parallel. Similar focusing across layers motivates a connection to the
central limit theorem: shared error across layers sets the loss plateau,
while aggregation turns layer-specific differences
into continued gains with depth. Our findings suggest that depth
scaling may arise from nonlinearity in attention, which allows large language
models to focus locally and may make the global covariance structure less
relevant.
\end{abstract}

\begin{figure}[H]
  \centering
  \includegraphics[width=0.74\linewidth]{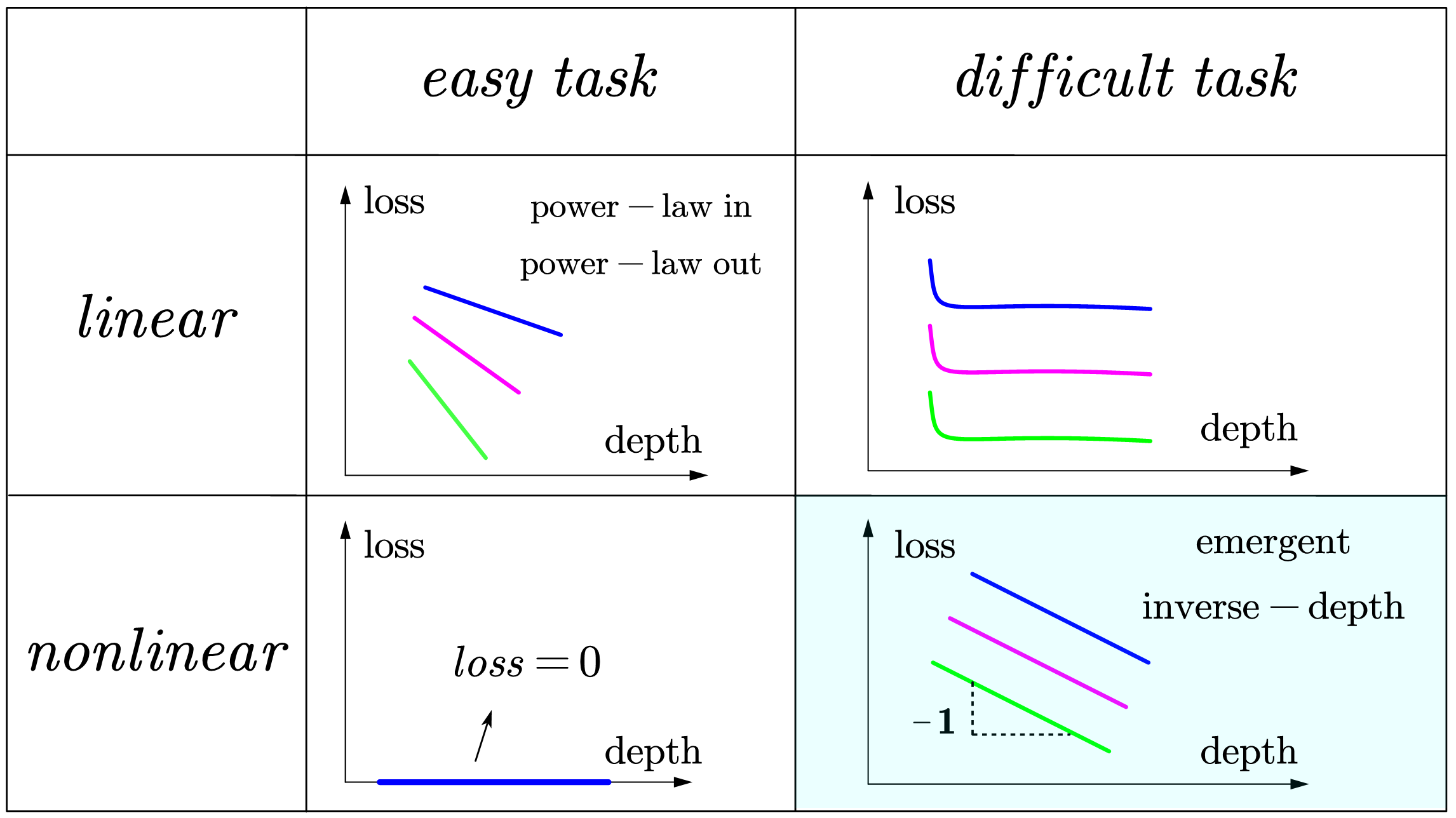}
  \caption{The nonlinear-attention model can yield inverse-depth scaling across all tested data spectra. This overview compares depth-dependent performance across models and tasks. 
  Colors distinguish data spectra, and all panels are plotted on a log-log scale.}
  \label{fig:overview}
\end{figure}

\section{Introduction}
\label{sec:introduction}

Neural scaling laws describe how model loss decreases with the size
of the dataset and the model, providing a practical basis for predicting performance and
allocating computational resources
\citep{hestness2017deep,kaplan2020scaling,hoffmann2022chinchilla}.
Despite their predictive success, these laws remain largely empirical, and the
mechanisms underlying their behavior are not fully understood. Clarifying their origins could
help establish the scope of these laws
and reveal how scaling can be made more efficient.

Recent theoretical work has begun to examine model-size scaling more closely
by separating the roles of width and depth
\citep{bahri2022explaining,liu2025superposition,bordelon2026theory}. Increasing width provides more representational capacity within
each layer, whereas increasing depth adds successive processing steps;
the two can therefore improve performance through different mechanisms.
Here, we focus on how models use additional layers and how this gives rise
to scaling with depth. Previous work on linear-attention models ties
power-law depth scaling to a power-law data spectrum
\citep{bordelon2026theory}. Yet it does not account for nonlinear attention in practical
large language models. In contrast, empirical work on these models suggests that
inverse-depth scaling can arise through an
ensemble of layers with similar behavior \citep{liu2026inverse}, without
explaining the underlying reason for this layerwise similarity. We therefore ask:

\begin{tcolorbox}[
  enhanced,
  colback=red!5,
  colframe=red!60!black,
  colbacktitle=red!60!black,
  coltitle=white,
  title={Our research question:},
  fonttitle=\bfseries,
  boxrule=0.8pt,
  arc=2pt,
  left=6pt,
  right=6pt,
  top=5pt,
  bottom=5pt,
  before skip=6pt,
  after skip=8pt
]
How does nonlinearity in attention affect depth scaling and its underlying
mechanisms?
\end{tcolorbox}

We study this question in the common setting of in-context learning (ICL),
where a model uses examples in its context to make predictions for a new
query. Our tasks model the second half of an induction head's computation:
retrieving the token that matches the query and copying its associated
value \citep{elhage2021framework,olsson2022induction,lv2025grok}. To isolate the role of nonlinearity, we
compare controlled Transformers with linear or nonlinear attention across
independently varied data spectra.

Figure~\ref{fig:overview} summarizes the resulting picture. In an easy task,
the linear-attention model follows a spectrum-dependent power law, while a
single-layer nonlinear-attention model is enough to solve the task. In a
difficult task, the linear-attention model quickly plateaus, whereas the
nonlinear-attention model exhibits inverse-depth loss scaling. Unlike the
linear-attention model, which cannot selectively focus on relevant tokens,
the nonlinear-attention model can attend to them directly, allowing strong
and weak spectral directions to be learned in parallel and enabling better
utilization of depth. Further analysis decomposes the loss into
cross-layer and within-layer terms and connects this decomposition to the
central limit theorem. The asymptotic
behavior of these terms explains the origins of the loss plateau and the
reducible loss.

Finally, we relax several simplifying assumptions of the controlled setup to
move closer to standard Transformers and observe similar signs of
inverse-depth scaling. These results suggest that nonlinearity in attention
can be a plausible source of depth scaling in real models and may make
scaling less dependent on the global covariance structure of the data.

\begin{tcolorbox}[
  enhanced,
  colback=red!5,
  colframe=red!60!black,
  colbacktitle=red!60!black,
  coltitle=white,
  title={Our contributions are the following:},
  fonttitle=\bfseries,
  boxrule=0.8pt,
  arc=2pt,
  left=6pt,
  right=6pt,
  top=4pt,
  bottom=4pt,
  before skip=7pt,
  after skip=8pt
]
\begin{list}{\textbullet}{%
  \setlength{\leftmargin}{18pt}%
  \setlength{\labelwidth}{6pt}%
  \setlength{\labelsep}{6pt}%
  \setlength{\itemsep}{1pt}%
  \setlength{\topsep}{0pt}%
  \setlength{\parskip}{0pt}%
  \setlength{\parsep}{0pt}%
  \setlength{\partopsep}{0pt}%
}
  \item We discover that nonlinearity in attention can qualitatively affect
  depth scaling, enabling emergent
  inverse-depth loss scaling across all tested spectra.
  \item We explain how nonlinear attention enables parallel learning of
  strong and weak spectral directions, in contrast to the sequential learning
  behavior of linear-attention models.
  \item We connect inverse-depth scaling to the central limit theorem
  through a loss decomposition that identifies the origins of the loss
  plateau and the gains from depth.
\end{list}
\end{tcolorbox}

\section{Problem Setup}
\label{sec:setup}

\subsection{Task Setup}
\label{sec:task-setup}

In-context learning is a common setting in which predictions rely on
interactions among context tokens. These interactions expose the structure
of the data, providing a useful setting for studying the origins of
depth scaling. We consider a controlled ICL
problem in which each context contains $P$ inputs
$\vx_i\in\mathbb R^d$ with labels generated by a teacher vector
$\mBeta\in\mathbb R^d$:
\begin{equation}
  y_i=\mBeta^\top\vx_i,
  \qquad i=1,\ldots,P.
  \label{eq:linear-regression-task}
\end{equation}
Let $\lambda_k$ and $\omega_k$ denote the eigenvalues of the input and teacher
covariances, respectively. To control input and task structure independently,
we vary these spectra as
\begin{equation}
  \lambda_k\propto k^{-a},
  \qquad
  \omega_k\propto k^{-b},
  \qquad k=1,\ldots,d.
  \label{eq:task-power-law-spectra}
\end{equation}
Here, $a,b\geq0$ are the input-spectrum and teacher-spectrum exponents,
respectively. They independently control how rapidly the corresponding
variances decay across feature directions. Setting either exponent to zero
gives an isotropic covariance for the corresponding distribution.

In natural language, different contexts may share similar features but
express them along different directions. Following
\citet{bordelon2026theory}, we model this variation by independently sampling
an orthogonal rotation matrix $O\in\mathbb R^{d\times d}$ for each context.
Conditioned on $O$, the input and teacher covariances are
\begin{equation}
  \begin{gathered}
  \Sigma=\operatorname{Cov}(\vx_i\mid O)
  =O\operatorname{diag}(\lambda_1,\ldots,\lambda_d)O^\top,\\
  \Omega=\operatorname{Cov}(\mBeta\mid O)
  =O\operatorname{diag}(\omega_1,\ldots,\omega_d)O^\top.
  \end{gathered}
  \label{eq:task-covariances}
\end{equation}
Conditioned on $O$, we sample the teacher vector independently of the
context and test inputs, which are drawn i.i.d.:
\begin{equation}
  \begin{gathered}
    \mBeta
    \sim\mathcal{N}\!\left(
      \bm{0},\Omega\right),\\
    \vx_1,\ldots,\vx_P,\vx_{\mathrm{test}}
    \overset{\mathrm{i.i.d.}}{\sim}\mathcal{N}\!\left(\bm{0},\Sigma\right).
  \end{gathered}
  \label{eq:context-sampling}
\end{equation}
\begin{wrapfigure}{r}{0.52\textwidth}
  \vspace{-0.8\baselineskip}
  \centering
  \includegraphics[width=\linewidth]{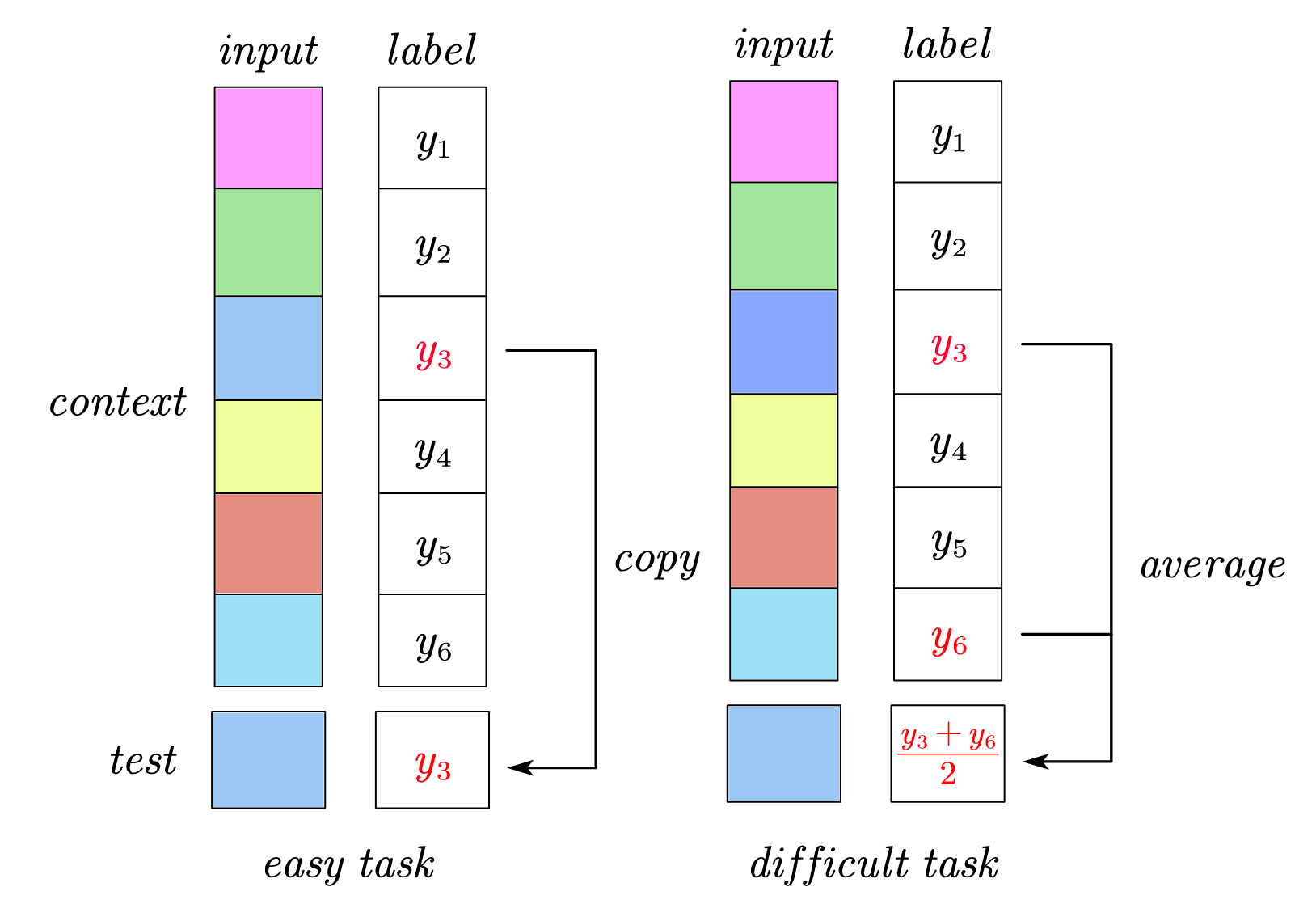}
  \caption{Both tasks model the second half of an induction head: the easy task copies the most relevant token’s label, while the difficult task averages the labels of the top-$k$ relevant tokens.}
  \label{fig:task-explanation}
  \vspace{0.4\baselineskip}
\end{wrapfigure}

The model must therefore infer the relevant directions from each context
rather than rely on fixed directions across contexts.

We first construct an \emph{easy task} using a widely studied ICL setting
\citep{garg2022what,vonoswald2023transformers,ahn2023preconditioned,bordelon2026theory}.
Given context inputs and their teacher-generated labels, the model must
predict the label assigned to a test input by the same teacher.
Following \citet{bordelon2026theory}, we consider the infinite-context limit
$P\to\infty$, in which a context token can be found arbitrarily close to
the test token. The task can therefore be approximated by retrieving the
closest context token and copying its label. This retrieval-and-copying
operation corresponds to the second half of an induction-head circuit
identified in language models
\citep{elhage2021framework,olsson2022induction,lv2025grok}.
The task can thus be viewed, in an idealized sense, as an exact-match problem:
\begin{equation}
  \vx_{\mathrm{test}}=\vx_j,
  \qquad y^*_{\mathrm{test}}=y_j,
  \qquad j\in\{1,\ldots,P\},\quad P\to\infty.
  \label{eq:easy-task-target}
\end{equation}

However, in practical ICL and language settings, prediction often requires combining information
from multiple context tokens rather than retrieving a single one. Motivated
by this need, we construct a \emph{difficult task} that extends
this induction-head-inspired operation from top-$1$ retrieval to top-$k$
aggregation. We measure relevance by cosine similarity and average the
selected labels. The selected index set and target are defined by
\begin{equation}
  \begin{aligned}
    S_k(\vx_{\mathrm{test}})
    &=\operatorname{TopK}_{i\in\{1,\ldots,P\}}
      \frac{\vx_i^\top\vx_{\mathrm{test}}}
      {\|\vx_i\|\,\|\vx_{\mathrm{test}}\|},\\
    y^*_{\mathrm{test}}
    &=\frac{1}{k}\sum_{i\in S_k(\vx_{\mathrm{test}})}y_i.
  \end{aligned}
  \label{eq:difficult-task-target}
\end{equation}
Here, $\operatorname{TopK}$ returns the indices of the $k$ largest scores. Figure~\ref{fig:task-explanation} summarizes these induction-head-inspired
top-$1$ copying and top-$k$ aggregation tasks.

\begin{figure}[t]
  \centering
  \begin{minipage}[t]{0.33\linewidth}
    \centering
    \begin{minipage}[b][1.34in][b]{\linewidth}
      \centering
      \includegraphics[width=\linewidth]{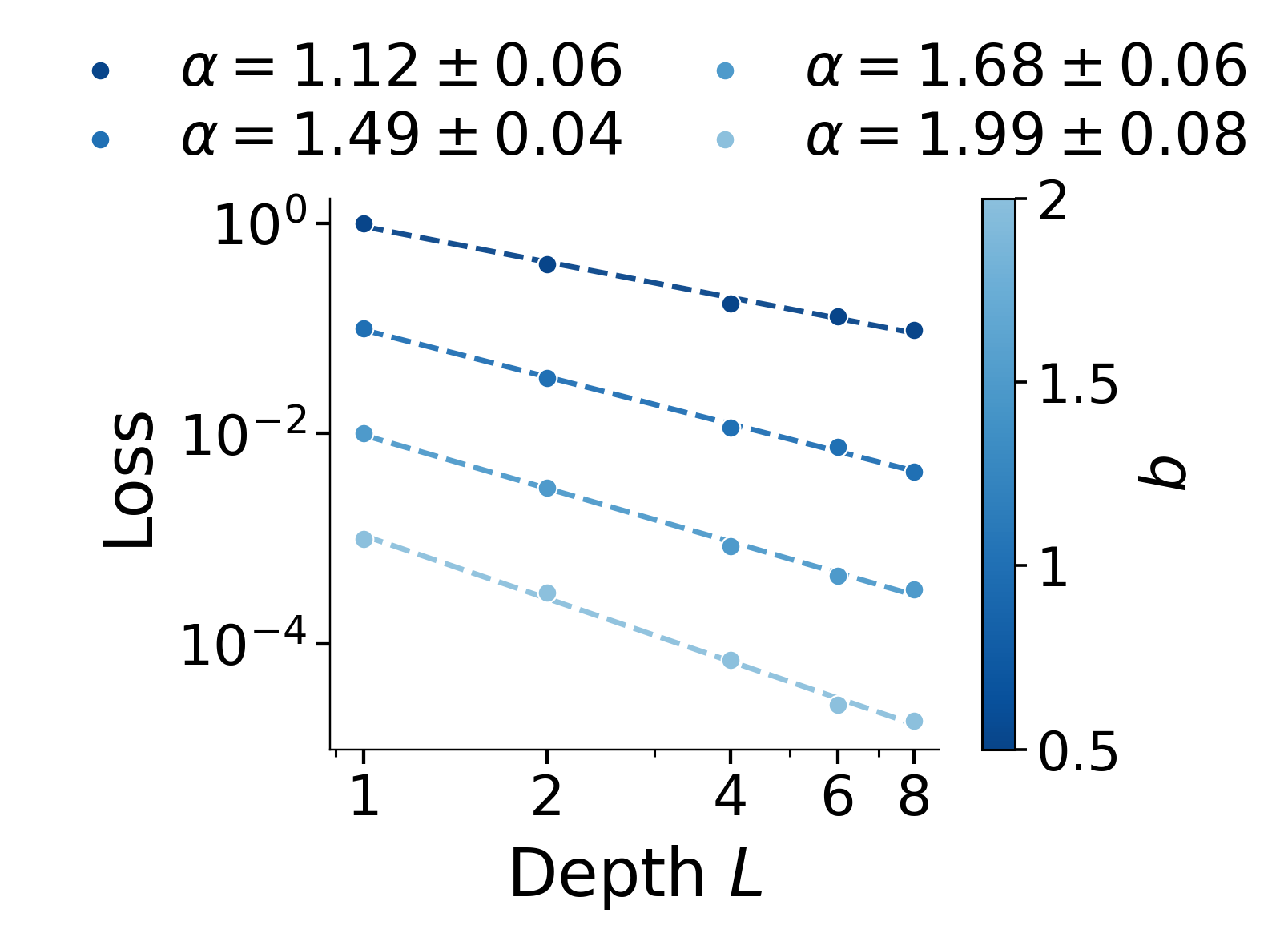}
    \end{minipage}
    \par\vspace{-1mm}
    {(a) Fixed $a=2$}
  \end{minipage}\hfill
  \begin{minipage}[t]{0.33\linewidth}
    \centering
    \begin{minipage}[b][1.34in][b]{\linewidth}
      \centering
      \includegraphics[width=\linewidth]{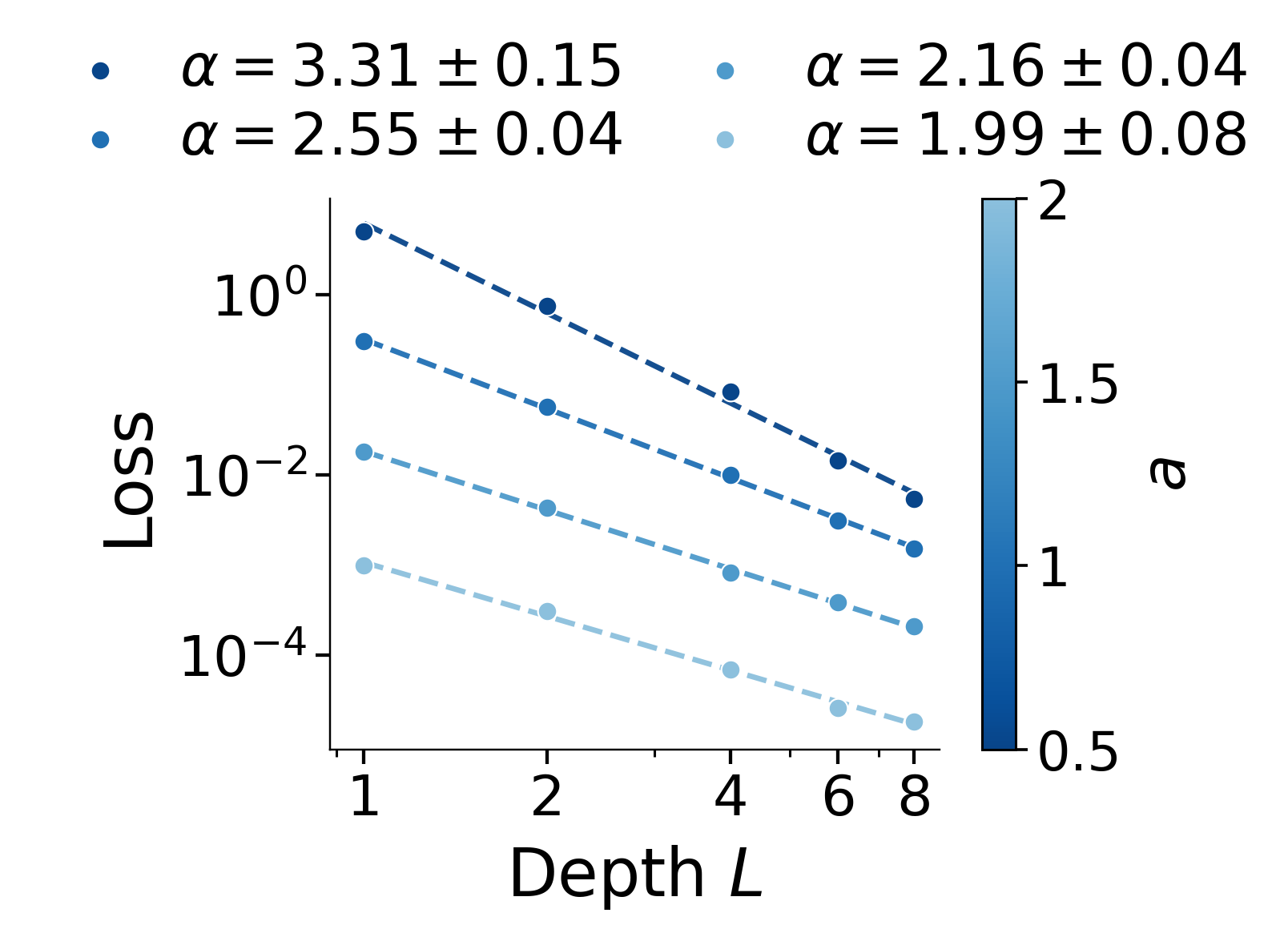}
    \end{minipage}
    \par\vspace{-1mm}
    {(b) Fixed $b=2$}
  \end{minipage}\hfill
  \begin{minipage}[t]{0.33\linewidth}
    \centering
    \begin{minipage}[b][1.34in][b]{\linewidth}
      \centering
      \includegraphics[width=0.96\linewidth]{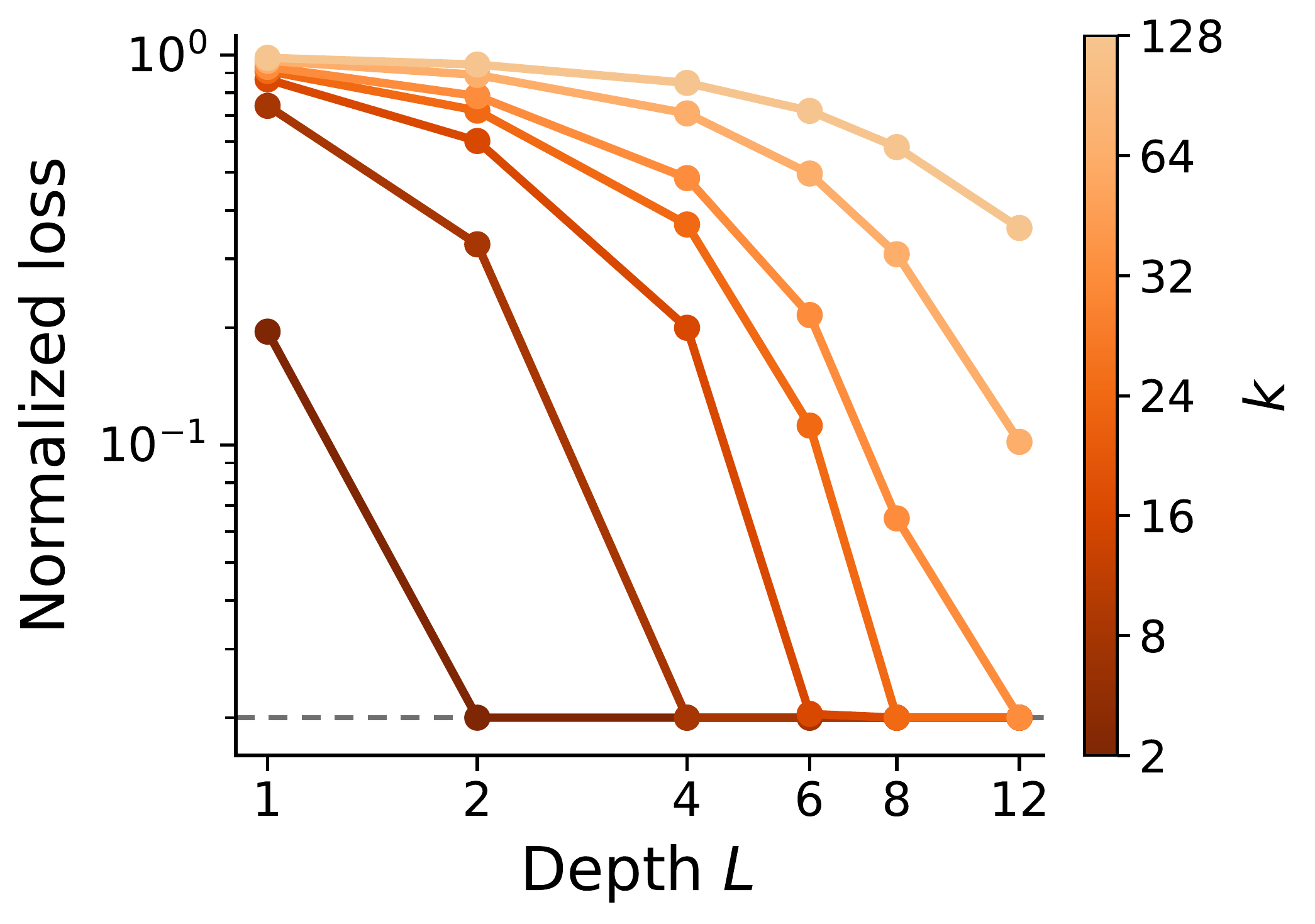}
    \end{minipage}
    \par\vspace{-1mm}
    {(c) Normalized mode-wise loss}
  \end{minipage}
  \caption{On the easy task, the linear-attention model exhibits spectrum-dependent
  depth scaling and learns spectral modes sequentially.
  (a) Varying $b$ at fixed $a=2$. (b) Varying $a$ at fixed $b=2$.
  Dashed curves show power-law fits, with the corresponding exponents indicated.
  (c) Normalized loss of representative spectral modes. Values below $2\%$
  are displayed at the threshold marked by the dashed line.}
  \label{fig:easy-linear-depth-scaling}
\end{figure}

\begin{figure}[t]
  \centering
  \begin{minipage}[c]{0.325\linewidth}
    \centering
\includegraphics[width=\linewidth,trim=0.35in 0 0 0,clip]{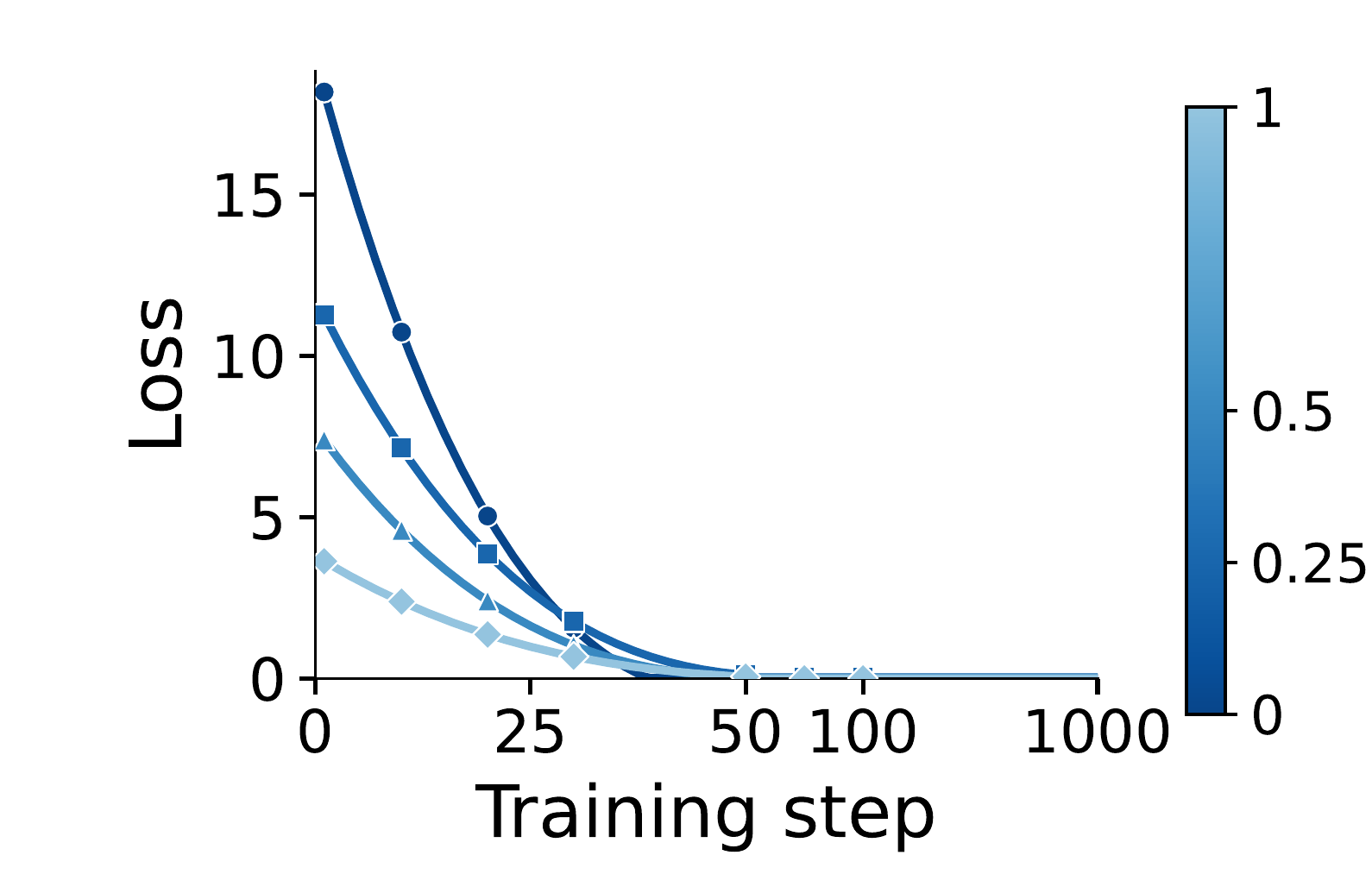}
    \vspace{-1mm}

    {\small (a) Fixed $b=0$}
  \end{minipage}\hfill
  \begin{minipage}[c]{0.325\linewidth}
    \centering
\includegraphics[width=\linewidth,trim=0.35in 0 0 0,clip]{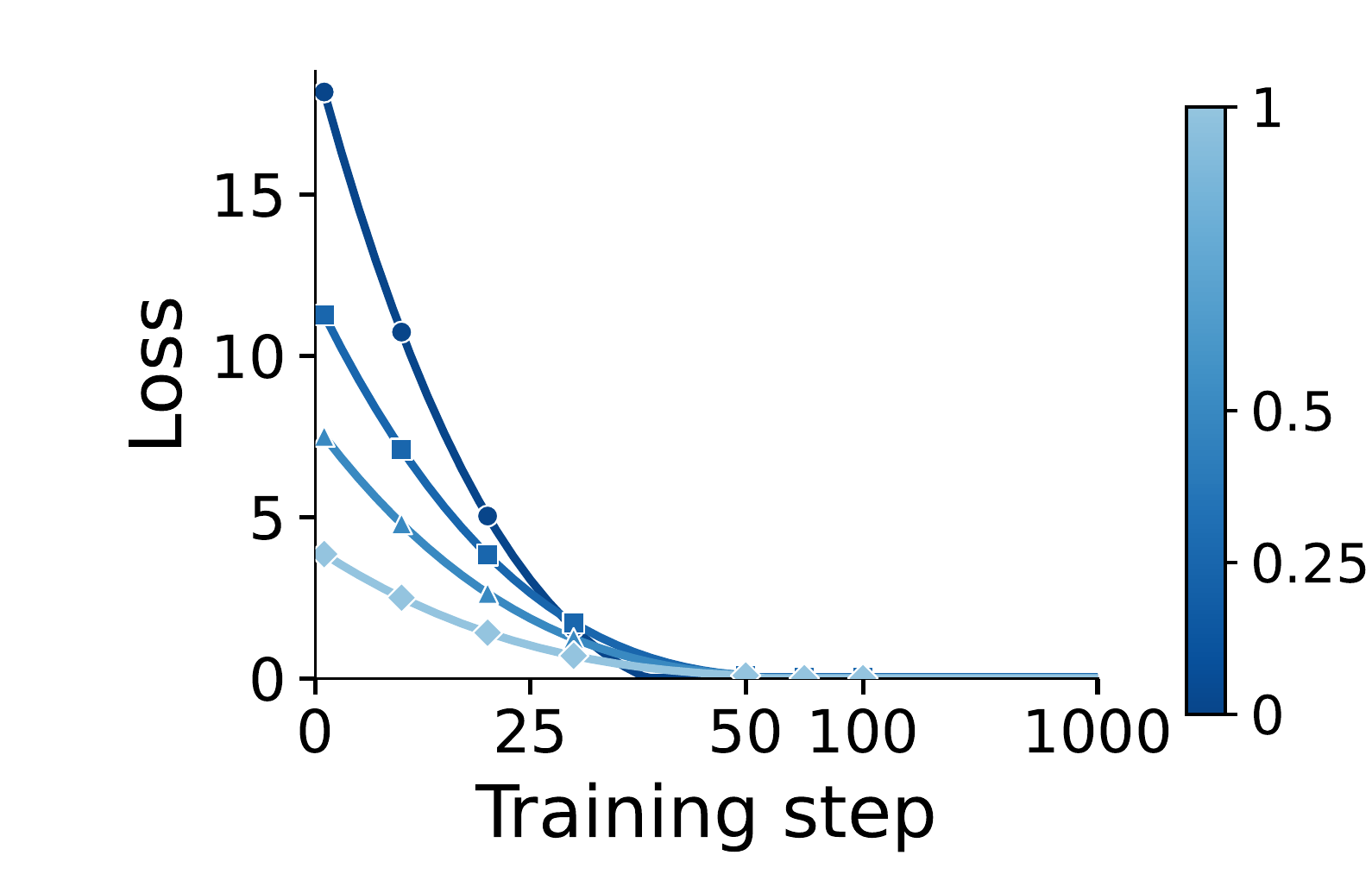}
    \vspace{-1mm}

    {\small (b) Fixed $a=0$}
  \end{minipage}\hfill
  \begin{minipage}[c]{0.325\linewidth}
    \centering
\includegraphics[width=\linewidth,trim=0.35in 0 0.35in 0,clip]{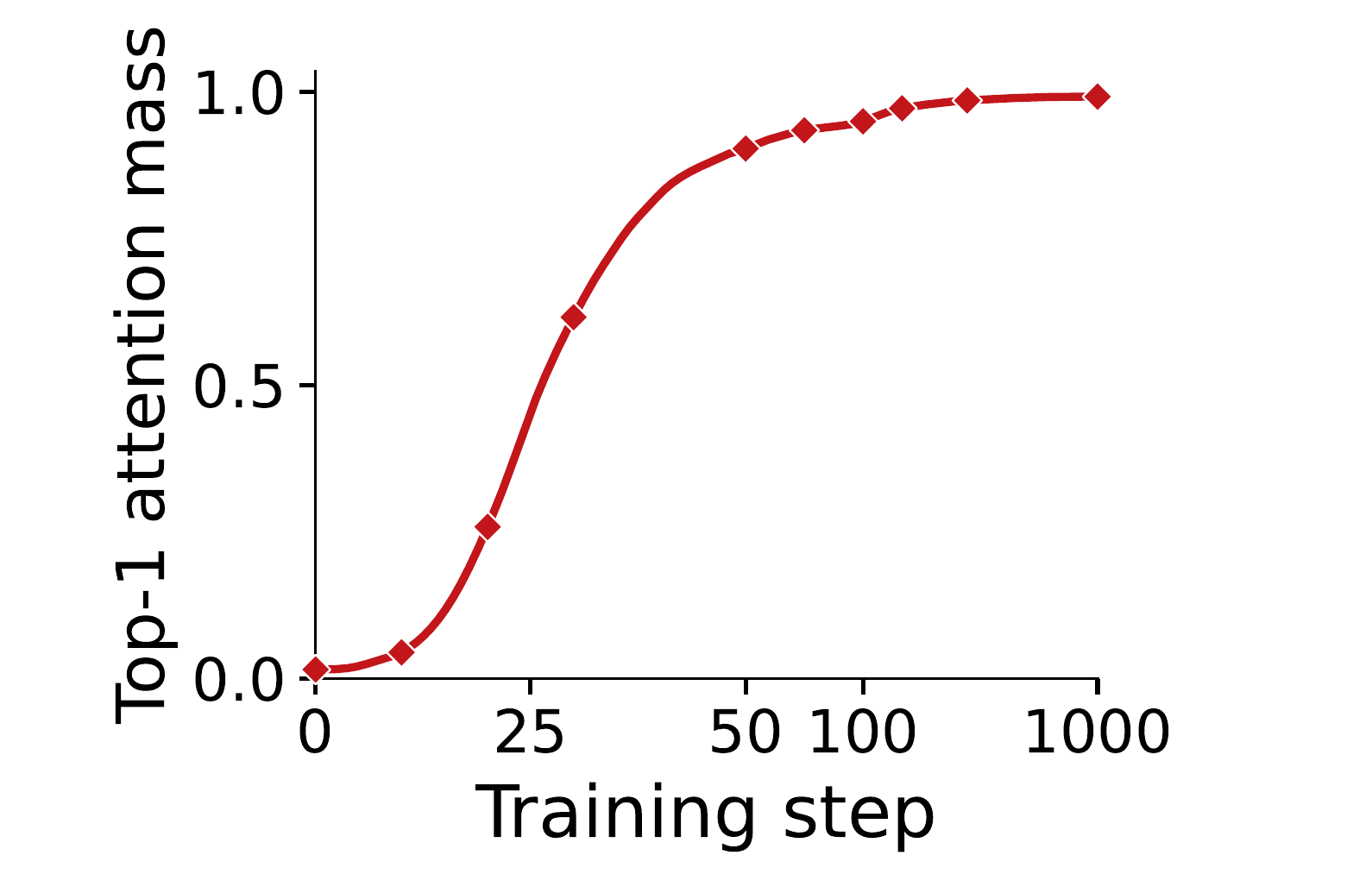}
    \vspace{-1mm}

    {\small (c) Top-$1$ attention mass, $(1,0)$}
  \end{minipage}
  \caption{A single nonlinear-attention layer solves the easy task by focusing
  on the matching token. (a) Varying $a$ at fixed $b=0$. (b) Varying $b$ at fixed $a=0$.
(c) Top-$1$ attention mass approaches one during training.}
  \label{fig:easy-task}
\end{figure}

\begin{figure}[t]
  \centering
  \begin{minipage}[t]{0.485\linewidth}
    \centering
\includegraphics[width=\linewidth]{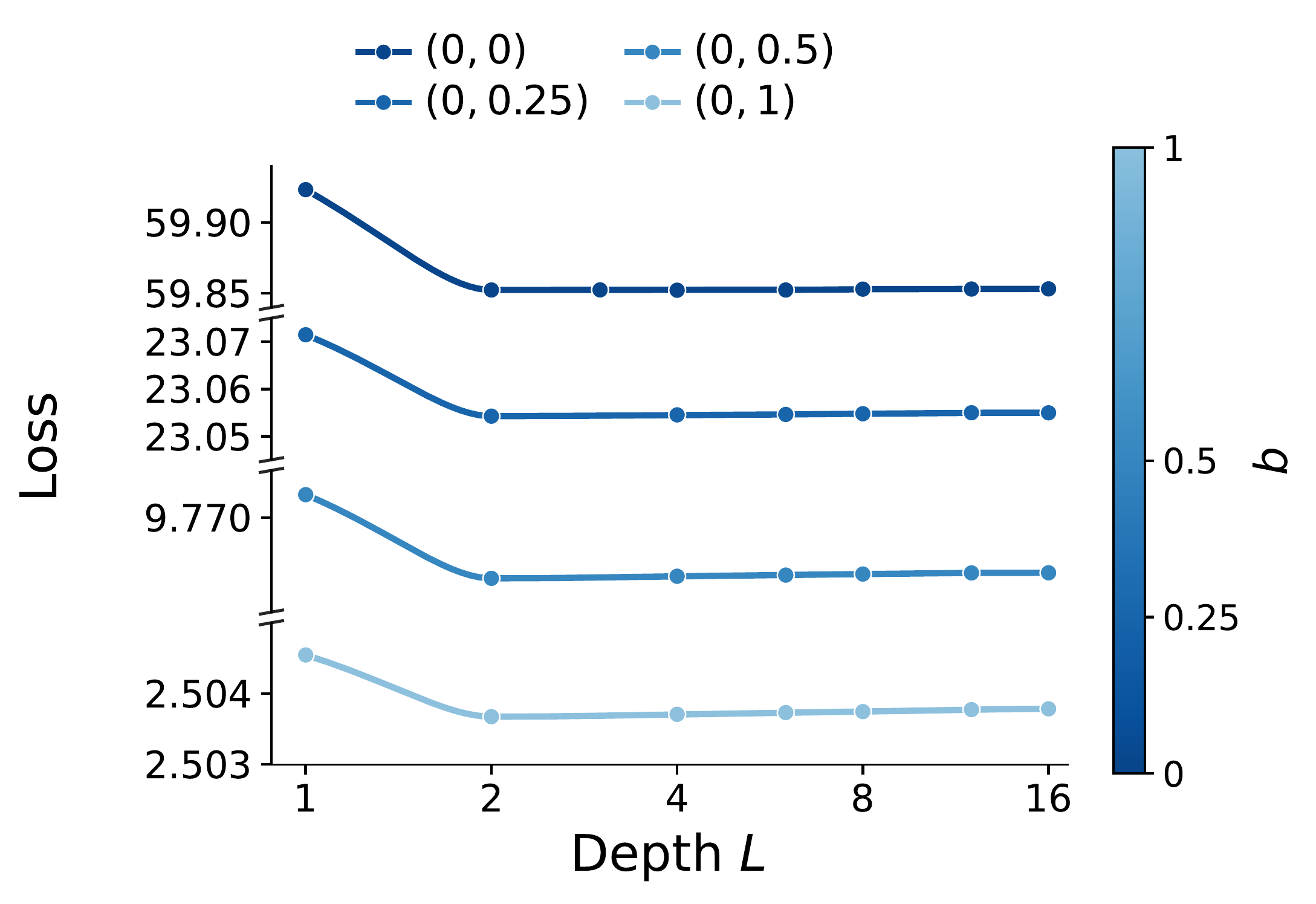}
    \vspace{-1mm}

    {(a) Fixed $a=0$}
  \end{minipage}\hfill
  \begin{minipage}[t]{0.485\linewidth}
    \centering
\includegraphics[width=\linewidth]{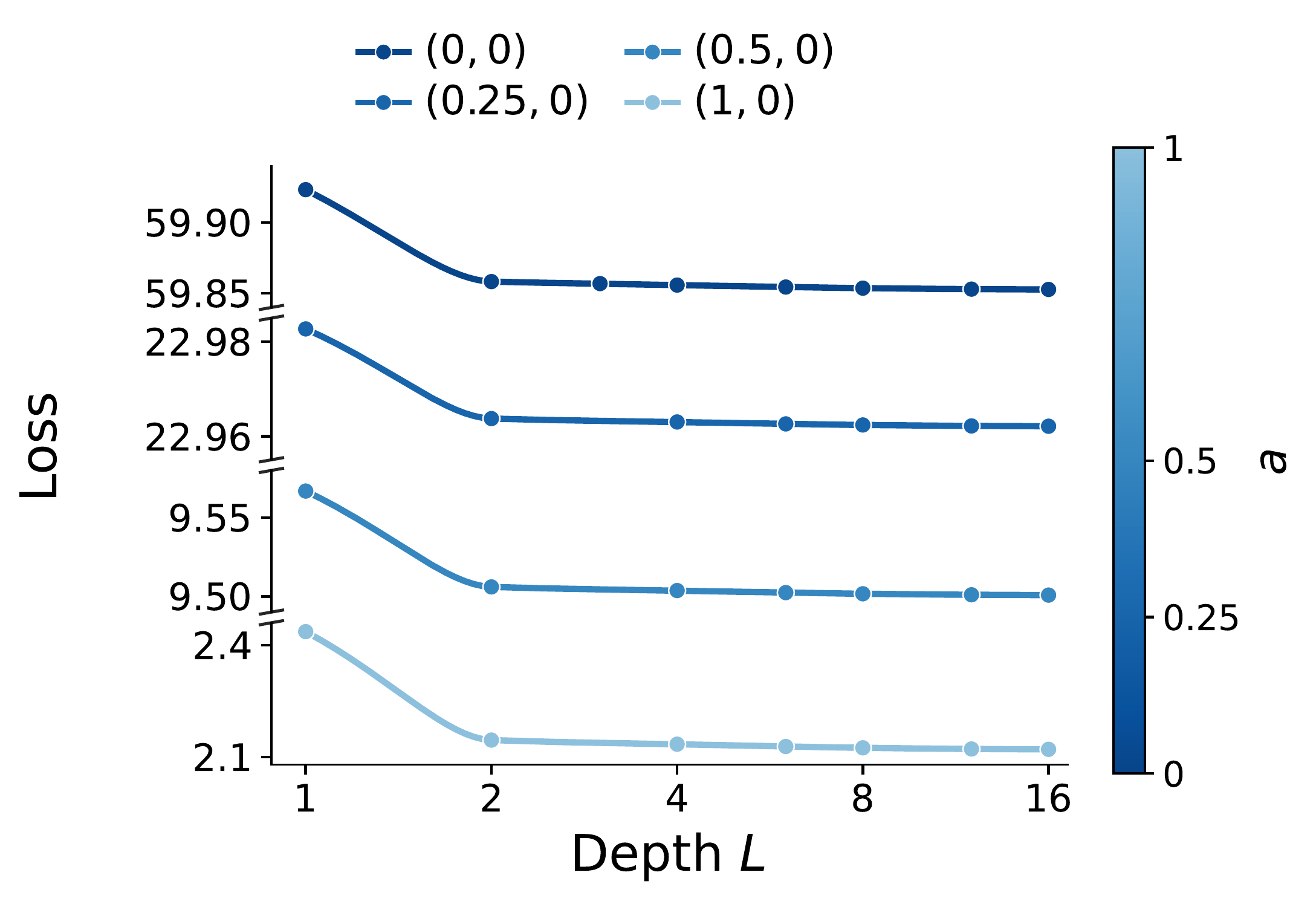}
    \vspace{-1mm}

    {(b) Fixed $b=0$}
  \end{minipage}
  \caption{On the difficult task, linear attention rapidly reaches a
  spectrum-dependent loss plateau with little benefit from further depth. (a) Varying $b$ at fixed $a=0$. (b) Varying $a$ at fixed $b=0$.}
  \label{fig:difficult-linear}
\end{figure}

\begin{figure}[t]
  \centering
  \begin{minipage}[t]{0.485\linewidth}
    \centering
\includegraphics[width=\linewidth]{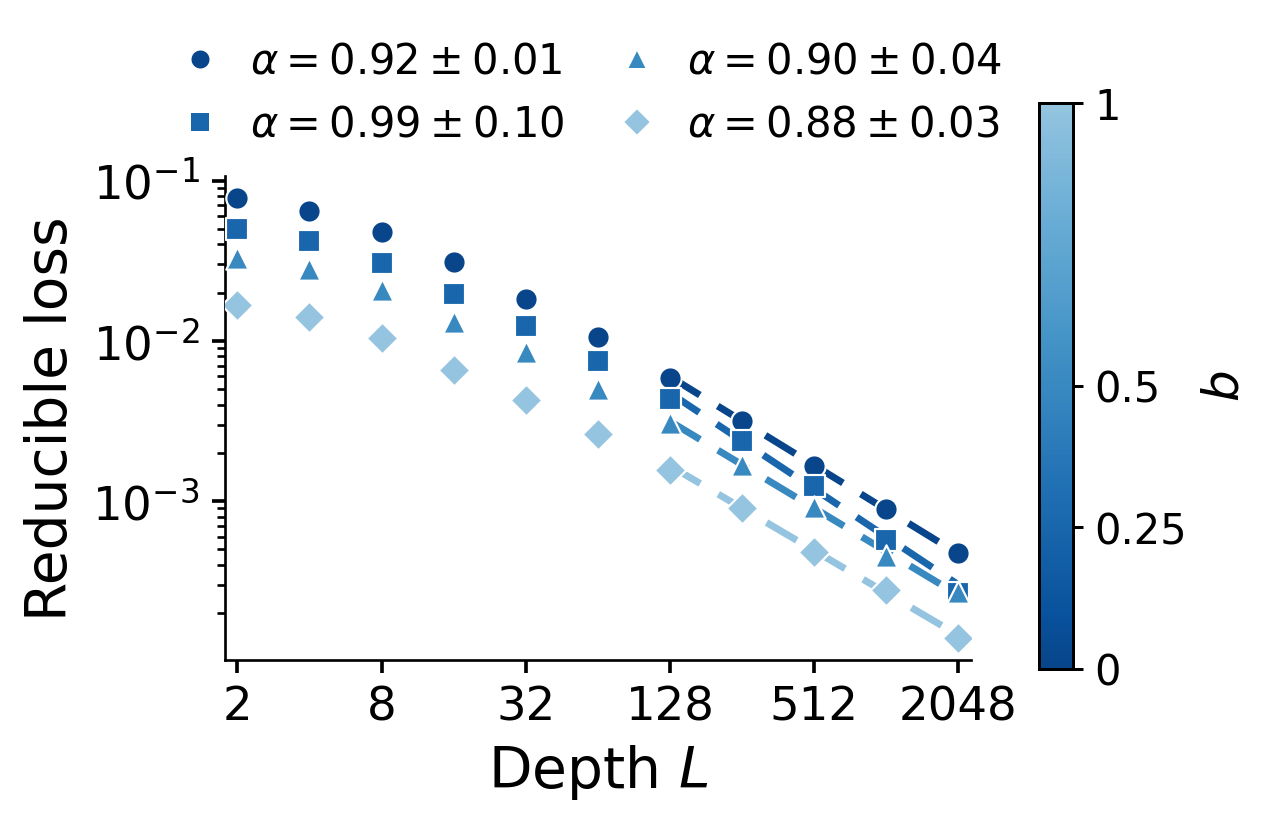}
    \vspace{-1mm}

    {\small (a) Fixed $a=0$}
  \end{minipage}\hfill
  \begin{minipage}[t]{0.485\linewidth}
    \centering
\includegraphics[width=\linewidth]{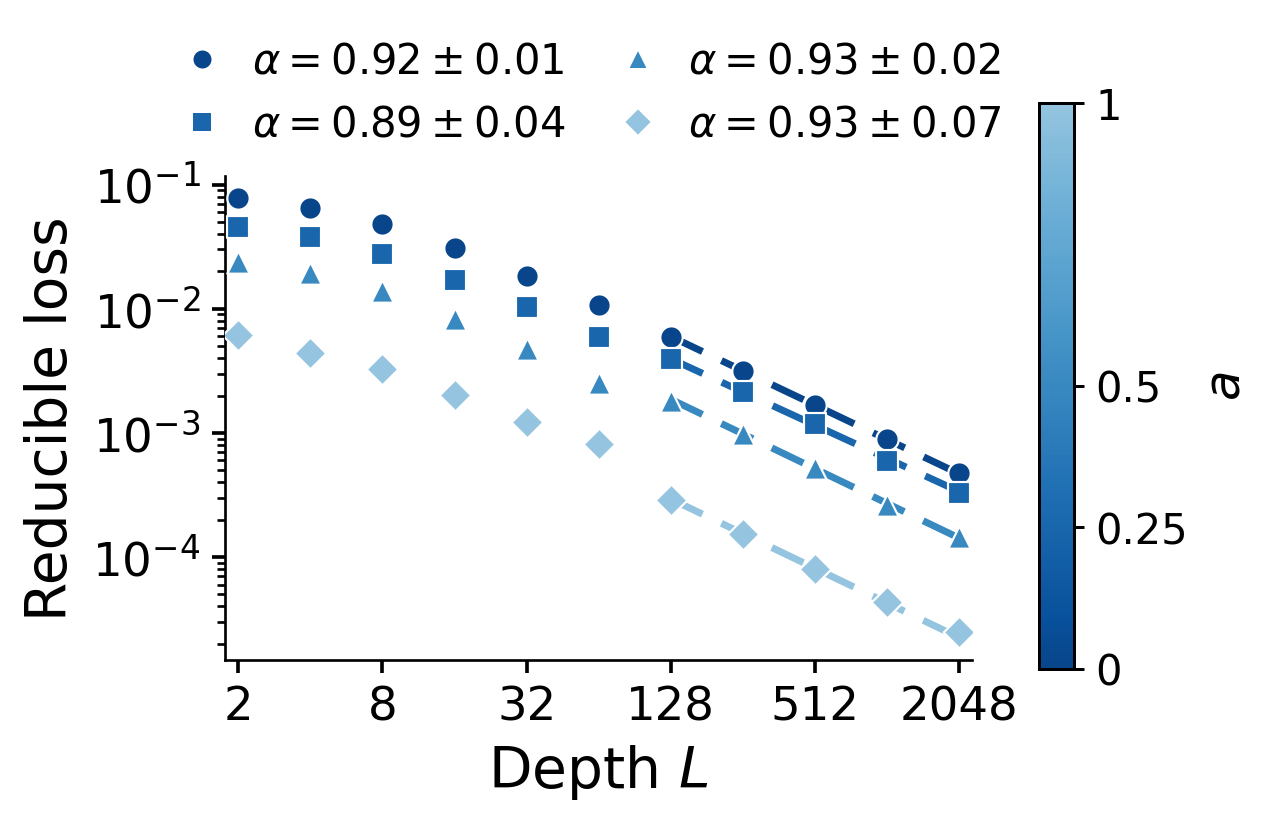}
    \vspace{-1mm}

    {\small (b) Fixed $b=0$}
  \end{minipage}
  \vspace{2mm}

  \begin{minipage}[t]{0.485\linewidth}
    \centering
\includegraphics[width=\linewidth]{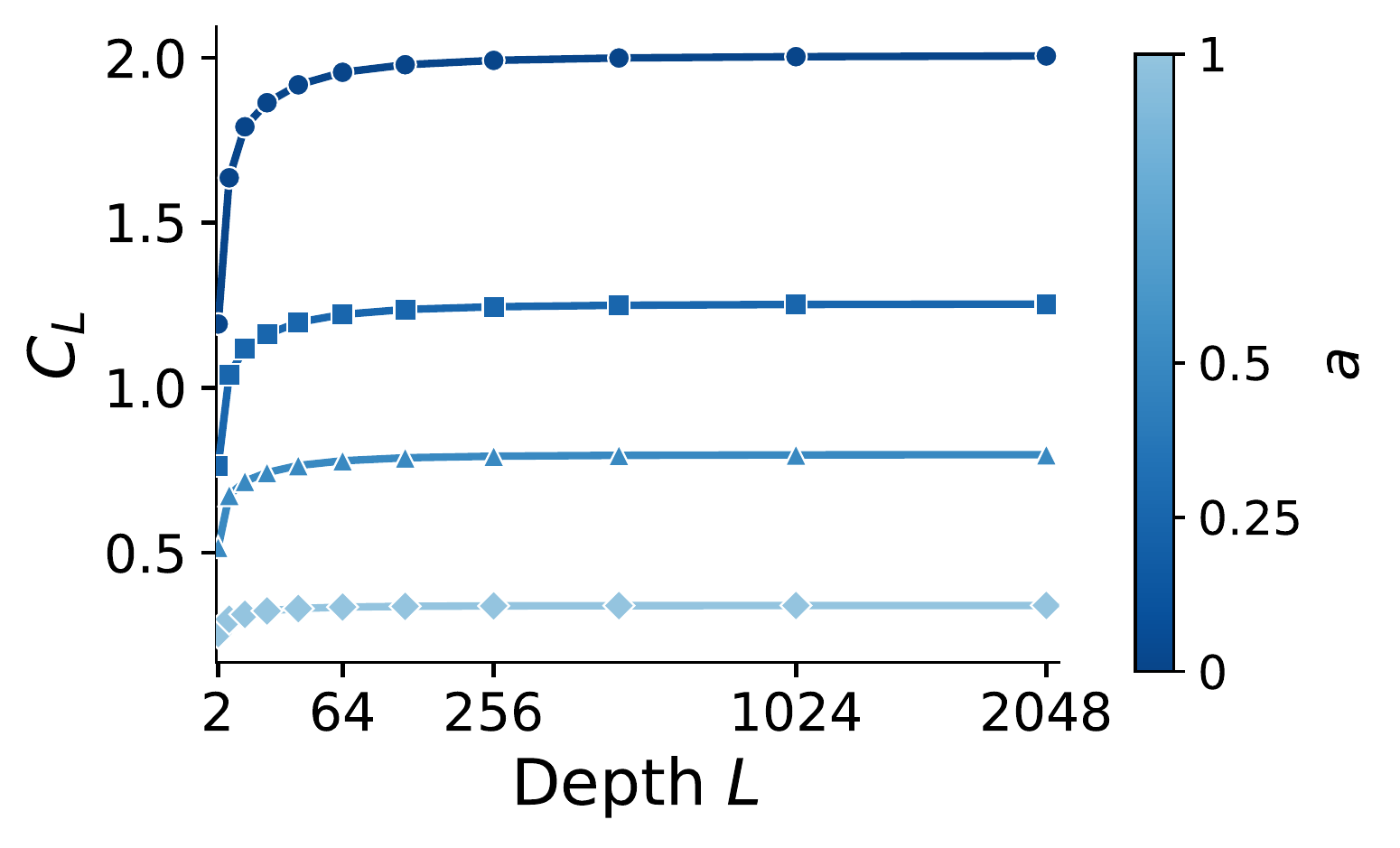}
    \vspace{-1mm}

    {\small (c) Shared-error term $C_L$ at fixed $b=0$}
  \end{minipage}\hfill
  \begin{minipage}[t]{0.485\linewidth}
    \centering
\includegraphics[width=\linewidth]{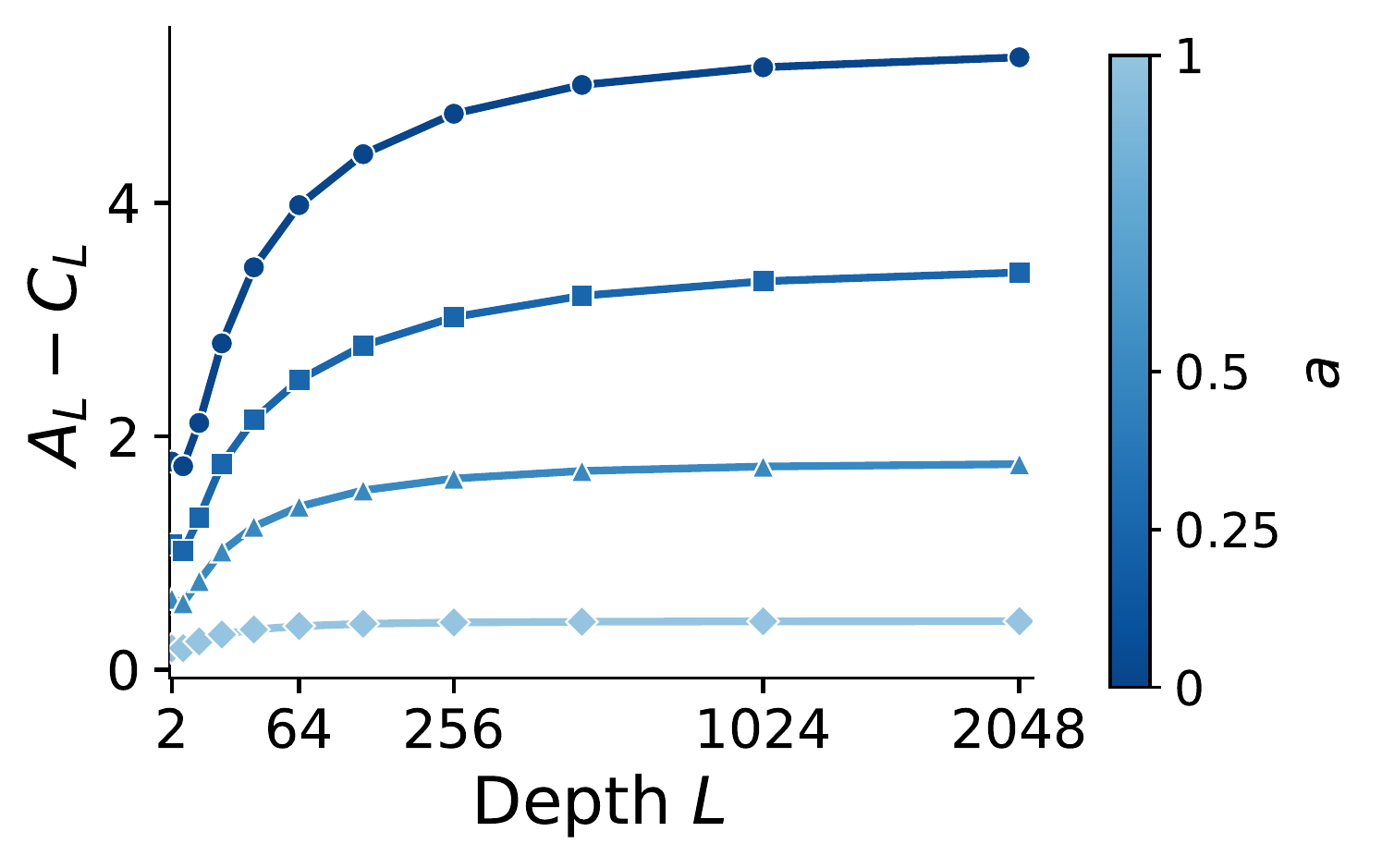}
    \vspace{-1mm}

    {\small (d) Disagreement term $A_L-C_L$ at fixed $b=0$}
  \end{minipage}
  \caption{On the difficult task, nonlinear attention yields
  inverse-depth decay across all tested data spectra. (a) Varying $b$ at fixed $a=0$. (b) Varying $a$ at fixed $b=0$.
(c,d) $C_L$ and $A_L-C_L$ converge to
spectrum-dependent constants.}
  \label{fig:difficult-nonlinear}
\end{figure}

\begin{figure}[t]
  \centering
  \begin{minipage}[t]{0.485\linewidth}
    \centering
\includegraphics[width=\linewidth]{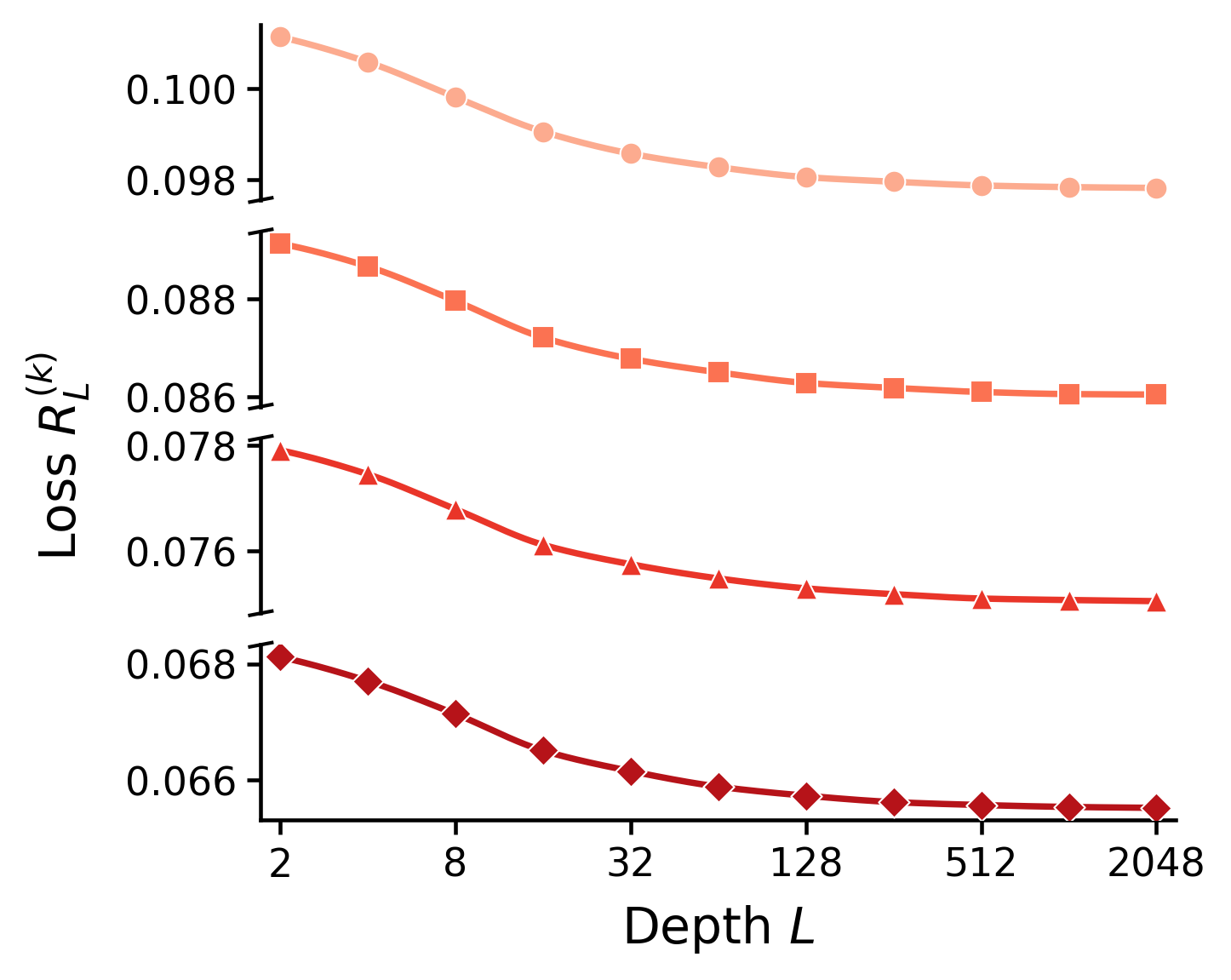}
    \vspace{-1mm}

    {\small (a) Mode-wise raw loss}
  \end{minipage}\hfill
  \begin{minipage}[t]{0.485\linewidth}
    \centering
\includegraphics[width=\linewidth]{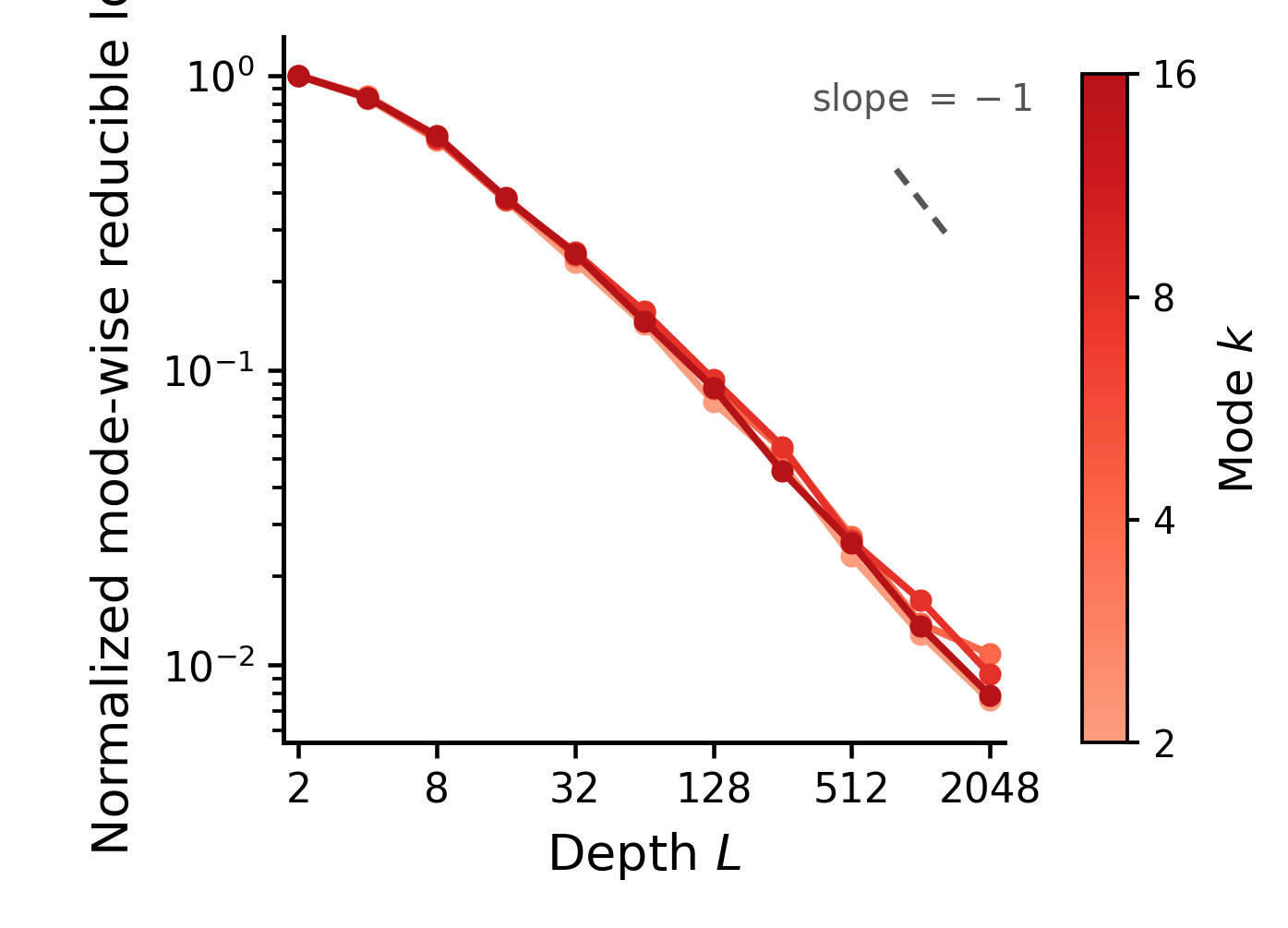}
    \vspace{-1mm}

    {\small (b) Normalized mode-wise reducible loss}
  \end{minipage}
  \caption{With nonlinear attention, strong and weak spectral modes are learned
  in parallel. (a) Different modes are learned at similar rates.
  (b) Normalized mode-wise reducible losses nearly collapse onto a common curve, indicating
  parallel learning across modes.}
  \label{fig:single-mode-scaling}
\end{figure}

\subsection{Model Setup}
\label{sec:model-setup}

Large language models combine positional encodings, multiple attention
heads, MLP blocks, normalization, and task-specific readouts. While important
in practice, these components make it difficult to isolate the mechanisms
underlying depth scaling. Building on \citet{bordelon2026theory}, we therefore
use a simplified architecture that retains the components needed for
context-dependent retrieval and label propagation, allowing a controlled
comparison between linear and nonlinear attention.

\paragraph{Embeddings.} Each context provides $P$ input--label pairs
$(\vx_i,y_i)$, where the teacher vector $\mBeta$ generates $y_i$ from $\vx_i$
according to \eqref{eq:linear-regression-task}. The model has access to all
context input--label pairs and the test input, but cannot access the test
label. The corresponding entry at the test position is initialized to zero.
We collect the token inputs and the
observed labels into
\begin{equation}
  X=[\vx_1,\ldots,\vx_P,\vx_{\mathrm{test}}]
  \in\mathbb R^{d\times(P+1)},
  \qquad
  \vy^{\mathrm{in}}=(y_1,\ldots,y_P,0)
  \in\mathbb R^{1\times(P+1)}.
  \label{eq:model-inputs}
\end{equation}
Let $N\geq d$ denote the hidden dimension, avoiding a representational
bottleneck. Each token state lies in $\mathbb R^N$. With encoder matrices
$W_x\in\mathbb R^{N\times d}$ and $W_y\in\mathbb R^{N\times1}$, we encode
input and label information into separate streams:
\begin{equation}
  H_x=W_xX\in\mathbb R^{N\times(P+1)},
  \qquad
  H_y^0=W_y\vy^{\mathrm{in}}\in\mathbb R^{N\times(P+1)}.
  \label{eq:two-stream-encoding}
\end{equation}
This encoding separates the input features used for token matching from
the labels to be propagated. The input features determine which context
tokens are relevant to the query, and attention passes their label
information to the test token.

The model has $L$ residual layers, indexed by $\ell\in\{1,\ldots,L\}$.
At each layer, attention uses the input features to determine how label
information is propagated. For simplicity, we first keep $H_x$ fixed
across layers, while $H_y^\ell$ is updated as label information is passed
among tokens. We use layer-specific
query and key matrices $W_q^\ell,W_k^\ell\in\mathbb R^{N\times N}$ and a
shared value matrix $W_v\in\mathbb R^{N\times N}$. Thus, each layer can
select relevant tokens differently while using a common transformation
to propagate label information.

\paragraph{Nonlinear attention.}
We first define the model with nonlinear attention. We apply pre-norm to the
input-stream hidden state for better training stability and alignment with
practical models, and denote the normalized state by $\widehat H_x$.
At layer $\ell$, the
query, key, and value representations are
\begin{equation}
  Q^\ell=W_q^\ell\widehat H_x,
  \qquad
  K^\ell=W_k^\ell\widehat H_x,
  \qquad
  V^\ell=W_vH_y^{\ell-1},
  \qquad
  Q^\ell,K^\ell,V^\ell\in\mathbb R^{N\times(P+1)}.
  \label{eq:nonlinear-qkv}
\end{equation}
Let $\mathsf M$ denote the attention-mask matrix. The attention matrix and residual
update are
\begin{equation}
  \begin{aligned}
    A^\ell
    &=\mathsf M\odot\operatorname{softmax}\!\left((Q^\ell)^\top K^\ell\right)
    \in\mathbb R^{(P+1)\times(P+1)},\\
    H_y^\ell
    &=H_y^{\ell-1}+\frac{1}{L}V^\ell
    (A^\ell)^\top
    \in\mathbb R^{N\times(P+1)}.
  \end{aligned}
  \label{eq:nonlinear-attention}
\end{equation}
\paragraph{Linear attention.}
The linear-attention model removes nonlinearities and retains the raw bilinear
attention scores. Its attention matrix and residual update are
\begin{equation}
  \begin{aligned}
    A^\ell
    &=\frac{1}{PN}\mathsf M\odot\left((Q^\ell)^\top K^\ell\right)
    \in\mathbb R^{(P+1)\times(P+1)},\\
    H_y^\ell
    &=H_y^{\ell-1}+\frac{1}{L}V^\ell
    (A^\ell)^\top
    \in\mathbb R^{N\times(P+1)}.
  \end{aligned}
  \label{eq:linear-attention}
\end{equation}

\paragraph{Attention masks.}
As in large language models, we use attention masks to control information
flow between tokens. For the linear-attention model, we adopt the masking
scheme of \citet{bordelon2026theory}, allowing successive layers to extract
and propagate information at different levels of granularity. For the nonlinear-attention model, we
initially disable context updates to focus on how different
layers retrieve information from a fixed context. In both models, the test
token can attend to all context tokens to access their label information.
Note that in the nonlinear-attention model, softmax is computed only over
tokens that can be attended to, so disabled positions do not affect the distribution.
Further details on the masks and context updates are given in
Appendix~\ref{app:context-updates}.

\paragraph{Output and loss.}
We obtain the prediction by applying a linear readout
$W_o\in\mathbb R^{1\times N}$ to the final test-token state
$h_{y,\mathrm{test}}^L$. The model is trained to minimize the mean squared
error over independently sampled contexts:
\begin{equation}
  \widehat y_L=W_oh_{y,\mathrm{test}}^L,
  \qquad
  \mathcal R_L
  =\mathbb E_{\mathrm{context}}
  \!\left[(\widehat y_L-y^*)^2\right].
  \label{eq:final-loss}
\end{equation}
Throughout, $\mathcal R_L$ denotes the evaluation mean squared error of a
depth-$L$ model; we write $\mathcal R_L(a,b)$ when making its dependence on
the data-spectrum exponents explicit.

All models are trained until the loss has converged. In
Appendix~\ref{app:generalized-models}, we relax
several simplifying assumptions to better reflect the architectures of
standard Transformers and large language models, and extend our
experiments and analyses to these settings.

\section{Easy Task Analysis}
\label{sec:easy-task-analysis}

\subsection{Linear-Attention Model}
\label{sec:easy-linear}

Figure~\ref{fig:easy-linear-depth-scaling}(a,b) shows that the
linear-attention model's loss on the easy task decreases with depth
according to a power law across all tested settings, with an exponent
that depends on the data spectra.

Heuristically, we can interpret the depth-$L$ linear-attention model's output as a
polynomial in the data covariance:
\begin{equation}
  \widehat y_L=\mBeta^\top p_L(\Sigma)\vx_{\mathrm{test}},
  \qquad
  p_L(\Sigma)=\sum_{r=1}^{L}a_r\Sigma^r.
  \label{eq:linear-polynomial-predictor}
\end{equation}
The target is
\begin{equation}
  y^*=\mBeta^\top\vx_{\mathrm{test}}.
  \label{eq:easy-identity-approximation}
\end{equation}
Therefore, the model needs to approximate the identity map with
$p_L(\Sigma)$; Appendix~\ref{app:linear-polynomial-theory} derives this
covariance-polynomial representation.
Greater depth allows the model to use higher powers of the covariance,
incorporating information at different levels of granularity to improve
the approximation. The asymptotic behavior of this approximation gives rise
to power-law depth scaling.
Different data spectra change the covariance $\Sigma$
in the polynomial, leading to different convergence rates with depth. This produces
spectrum-dependent depth scaling. 

This mechanism also has a feature-learning interpretation. We decompose the
power-law input covariance into individual spectral modes and normalize each
mode's loss, which allows us to compare how fully each
mode has been learned. Small $k$ indexes stronger features
with large eigenvalues, whereas large $k$ indexes weaker features with small
eigenvalues. Figure~\ref{fig:easy-linear-depth-scaling}(c) reveals a clear
separation in their learning dynamics: stronger modes are learned almost
perfectly within a few layers, while weaker modes require substantially greater
depth. Appendix~\ref{app:exponential-superposition} further shows that
individual modes can converge exponentially at different characteristic
depths, with their superposition yielding spectrum-dependent power-law scaling.

\subsection{Nonlinear-Attention Model}
\label{sec:easy-nonlinear}

We next evaluate the nonlinear-attention model on the easy task. Figure~\ref{fig:easy-task}(a,b) shows that a single-layer
nonlinear-attention model achieves zero loss on the easy task across all
tested data spectra. Unlike the spectrum-dependent improvement
of the linear-attention model, its performance does not improve further
with depth.

The key distinction is that nonlinear attention can selectively focus on
the relevant token in each context, whereas linear attention cannot fully
exclude contributions from irrelevant tokens. During training, the queries
and keys attain large norms, amplifying differences between attention
logits and placing softmax in an effectively low-temperature regime.
Almost all attention is then assigned to the matching token, allowing its
label to be copied in a single layer. Figure~\ref{fig:easy-task}(c) supports
this explanation: during training, attention shifts from a nearly uniform
distribution toward the top-$1$ token, whose attention mass approaches one. The same
qualitative behavior holds across the tested spectra.

\section{Difficult Task Analysis}
\label{sec:difficult-task-analysis}

\subsection{Linear-Attention Model}
\label{sec:difficult-linear}

We next evaluate the linear-attention model on the difficult task.
Here, we choose $k=2$ as a representative case of top-$k$ aggregation.
Figure~\ref{fig:difficult-linear}(a,b) shows that the loss reaches a
spectrum-dependent plateau within one or two layers. Unlike the continued
improvement on the easy task, additional depth provides little further
benefit.

Intuitively, the selected tokens vary across contexts, so their average
fluctuates around a shared mean. The target therefore contains a predictable
component and a context-specific deviation. Linear attention can capture
the shared component, but lacks the context-specific token selection needed
to fully recover the deviation. This leaves an irreducible loss that sets
the plateau.

We analyze the reducible part using the covariance-polynomial formulation
in Section~\ref{sec:easy-linear}. The prediction retains the same form,
while the predictable target has the leading-order form
\begin{equation}
  \widehat y_L=\mBeta^\top p_L(\Sigma)\vx_{\mathrm{test}},
  \qquad \overline{y}^{*}\propto\mBeta^\top\Sigma\vx_{\mathrm{test}},
  \label{eq:difficult-predictable-approximation}
\end{equation}
Unlike the identity map in the easy task, this leading target can already
be represented by a degree-one polynomial. Additional depth therefore
provides little further reduction in loss.
Appendix~\ref{app:difficult-linear-geometry} provides the geometric
illustration, detailed derivation, and comparison between predicted and
measured loss floors.

\subsection{Nonlinear-Attention Model}
\label{sec:difficult-nonlinear}

We next evaluate the nonlinear-attention model on the same difficult task
with $k=2$. Figure~\ref{fig:difficult-nonlinear}(a,b) shows that additional
depth continues to reduce the loss, in contrast to the rapid saturation
of the linear-attention model.

To quantify this depth dependence, we fit the evaluation loss to
\begin{equation}
  \mathcal R_L(a,b)=L_0(a,b)+c(a,b)L^{-\alpha(a,b)},
  \label{eq:free-exponent-fit}
\end{equation}
by minimizing squared residuals between the logarithms of the observed and
modeled losses. The fitted exponents shown in
Figure~\ref{fig:difficult-nonlinear}(a,b) are close to one across the tested
data spectra, indicating emergent inverse-depth scaling in the deep regime.

With input and label information carried in separate streams and context
states fixed across layers, our model admits an equivalent parallel
representation: each layer forms a prediction from the same context, and
the final output averages these predictions. Increasing depth therefore
increases the number of predictors in the ensemble, suggesting an analogy
with the central limit theorem, where averaging layer-specific fluctuations can
yield a $1/L$ decay in squared error.
To distinguish the error reduced by ensembling from the error that remains,
we derive the exact loss decomposition
\begin{equation}
  \mathcal R_L(a,b)
  =C_L(a,b)+\frac{A_L(a,b)-C_L(a,b)}{L}.
  \label{eq:finite-depth-decomposition}
\end{equation}
Here, $C_L$ measures error shared across predictors, whereas $A_L-C_L$
measures pairwise differences between their errors. A detailed derivation
is provided in Appendix~\ref{app:theory}.
Figure~\ref{fig:difficult-nonlinear}(c,d) tracks these two terms across
different data spectra. As depth increases, both converge to
spectrum-dependent constants:
\begin{equation}
  C_L(a,b)\longrightarrow C_\infty(a,b),
  \qquad
  A_L(a,b)-C_L(a,b)\longrightarrow\Delta_\infty(a,b),
  \qquad L\to\infty.
  \label{eq:nonlinear-error-limits}
\end{equation}

The loss plateau and the depth-dependent improvement have different
origins. The error shared across predictors cannot be
canceled by ensembling and sets the irreducible loss within this ensemble.
For the top-$2$ task, this shared error reflects the mismatch between
smooth softmax attention and the sharp target distribution
$(1/2,1/2,0,\ldots,0)$, which is difficult to match consistently across
contexts.

In contrast, $A_L-C_L$ quantifies error variation across predictors.
Ensembling allows these errors to partially cancel,
reducing their contribution by a factor of $1/L$ and enabling continued
improvement with depth.
Detailed derivations and attention-mass diagnostics
are provided in Appendices~\ref{app:theory}
and~\ref{app:attention-percentiles}, respectively.

We again examine individual spectral modes and normalize their excess losses
to compare relative learning progress. Figure~\ref{fig:single-mode-scaling}(a)
shows similar improvement rates across modes, and the normalized curves in
panel (b) nearly collapse. In contrast to the linear-attention model's
sequential learning, nonlinear attention can focus directly on relevant
tokens in each context and improve strong and weak features in parallel,
even without context updates that expose progressively finer-grained
information.

Building on our analysis of the differences between linear and nonlinear
attention, we further compare their irreducible losses on the difficult
task. Table~\ref{tab:main-nonlinear-fits} shows that the nonlinear-attention
model achieves a substantially lower loss floor, clearly outperforming
the linear-attention model.

\begin{table}[t]
  \centering
  \caption{Nonlinear-attention models achieve substantially lower irreducible
  loss than linear-attention models on the difficult task.}
  \label{tab:main-nonlinear-fits}
  \vspace{\baselineskip}
  \small
  \setlength{\tabcolsep}{4.5pt}
  \begin{tabular}{c*{7}{c}}
    \toprule
    $(a,b)$ & $(0,0)$ & $(0,0.25)$ & $(0,0.5)$ & $(0,1)$ & $(0.25,0)$ & $(0.5,0)$ & $(1,0)$ \\
    \midrule
    Nonlinear $L_0$ & 2.007 & 1.268 & 0.836 & 0.424 & 1.255 & 0.798 & 0.341 \\
    Linear $L_0$ & 59.853 & 23.055 & 9.766 & 2.504 & 22.962 & 9.501 & 2.120 \\
    \bottomrule
  \end{tabular}
\end{table}

We extend our setup toward standard Transformers by encoding inputs and labels
in a shared space, untying the value matrix, and updating context states across
layers. We observe similar signs of inverse-depth scaling
(Appendix~\ref{app:generalized-models}). These results suggest that this mechanism may
also contribute to depth scaling in realistic models with nonlinear
attention, where data structure need not be the dominant factor.

\section{Related Work}
\label{sec:related-work}

Empirical neural scaling laws relate loss to model size and dataset size, supporting performance prediction and resource allocation
\citep{hestness2017deep,kaplan2020scaling,hoffmann2022chinchilla}.
Theoretical explanations have related them to data geometry and covariance
spectra \citep{sharma2022manifold,bahri2022explaining}.
More recent work identifies mechanisms arising from representation
superposition and the dynamics of learning peaked distributions
\citep{liu2025superposition,liu2026onethird,kuehn2026boundary}.
Despite these advances, much of the literature treats model size as an
aggregate quantity rather than resolving the different contributions of
width and depth.

As a key component of model-size scaling, depth has been studied through
several complementary accounts of how networks utilize layers.
Compositional learning builds complex representations by successively
combining simpler features, making hierarchical structure more efficiently
learnable \citep{bengio2013representation,lecun2015deep,poggio2017why,cagnetta2024compositional}.
The dynamical-systems perspective treats residual layers as discrete steps
of an ordinary differential equation
\citep{e2017dynamical,haber2018stable,chen2018neural,sander2022discretize,chizat2025hidden}.
The ensemble-averaging perspective views depth as the aggregation of
related predictors
\citep{veit2016residual,lu2020meanfield,liu2026inverse}.
These perspectives explain how networks use depth but do not fully explain
the origins of depth scaling. We examine how nonlinearity in attention
shapes its underlying mechanisms and resulting scaling laws.

More directly, \citet{bordelon2026theory} build on the established ICL
regression setting \citep{garg2022what,vonoswald2023transformers} to derive
spectrum-dependent depth scaling for deep linear-attention models. A line of
theoretical work has largely focused on analytically tractable
linear models
\citep{ahn2023preconditioned,zhang2024trained,lu2025asymptotic}, leaving the
mechanisms through which nonlinear attention affects depth scaling unresolved.
\citet{liu2026inverse} report inverse-depth scaling in LLMs
and connect it to ensemble averaging, without explaining why this
ensemble-like behavior emerges. \citet{liu2026universality} further argue
that generic mechanisms may determine scaling exponents, while data and
architecture affect the coefficients.
Our work complements these studies
by explaining how nonlinear attention enables ensemble-like behavior
across layers and gives rise to inverse-depth scaling.

\section{Discussion}
\label{sec:discussion}

Our results identify nonlinearity in attention as a source of depth scaling
beyond power-law data spectra.
Nonlinear attention can focus directly on relevant tokens in each context,
allowing strong and weak spectral directions to be learned in parallel.
Linear attention lacks this selective ability, tying its learning dynamics
to global spectral strength.
Similar focusing across layers supports an ensemble interpretation:
aggregation reduces layer-specific errors while leaving a shared loss
floor, connecting inverse-depth improvement to the central limit theorem.
These findings suggest that practical depth scaling
can originate from attention nonlinearity rather than data structure alone.

Our study has many limitations.
Our simplified architecture omits components of full LLMs, so the
identified mechanism is not guaranteed to operate in those models.
Although architectural extensions show similar scaling signals, a clear
theoretical account of their mechanisms is still lacking.
Our experiments are restricted to ICL tasks, leaving
generalization to real-world datasets and broader task classes untested.
The identified mechanism need not be universal or dominant, and other
processes may produce different scaling or interact with it.
Our analysis also does not address how depth scaling couples to model width
and dataset size.

Aggregation also raises a question about how efficiently depth is used.
If a few layers perform the essential computation while others mainly
average residual errors, separate parameters at every layer may be wasteful.
Looped Transformers reuse parameters across depth
\citep{dehghani2019universal,geiping2025recurrent} and may retain
aggregation benefits with fewer parameters. Testing this possibility could
guide more parameter-efficient use of depth in future language models.

\subsection*{AI Use Statement}
We used generative AI tools to support brainstorming, articulate candidate
mathematical claims, generate code under explicit human instructions, and
assist with language editing. We did not use generative AI to generate
synthetic datasets, conduct data analysis, or write mathematical proofs.
All AI-assisted ideas and writing were critically reviewed by the authors,
and all AI-generated code was independently inspected and verified. The
authors performed the data analysis and mathematical proofs and take full
responsibility for the final content of this work.

\bibliography{references}
\bibliographystyle{preprint_style}

\clearpage
\appendix

\setcounter{topnumber}{2}
\setcounter{bottomnumber}{1}
\setcounter{totalnumber}{3}
\renewcommand{\topfraction}{0.90}
\renewcommand{\bottomfraction}{0.30}
\renewcommand{\textfraction}{0.05}
\renewcommand{\floatpagefraction}{0.70}
\setlength{\intextsep}{8pt plus 2pt minus 2pt}
\allowdisplaybreaks[2]

\section{Linear-Attention Model Theory}
\label{app:linear-theory}

\subsection{Covariance Polynomials in the Linear-Attention Model}
\label{app:linear-polynomial-theory}

We first derive the theory of the linear-attention model on the easy task. In general, its
output is an ordered matrix polynomial of degree at most $L$ in the data
covariance. The isotropic setting analyzed by \citet{bordelon2026theory}
then follows as a special case of our theory.

Let $U$ collect the context inputs, $q$ denote the query, and $S$ be the
empirical input covariance:
\begin{equation}
 \begin{gathered}
 U=[\vx_1,\ldots,\vx_P]\in\mathbb R^{d\times P},
 \qquad q=\vx_{\mathrm{test}}\in\mathbb R^d,\\
 y=U^\top\mBeta,\qquad S=\frac1P UU^\top.
 \end{gathered}
 \label{eq:app-rotated-covariances}
\end{equation}
The input and teacher covariances are
\[
 \begin{gathered}
 \Sigma=O\operatorname{diag}(\lambda_1,\ldots,\lambda_d)O^\top,\\
 \Omega=O\operatorname{diag}(\omega_1,\ldots,\omega_d)O^\top,
 \end{gathered}
\]
where the random orthogonal rotation $O$ is fixed within a context and
sampled independently across contexts. Contexts therefore share the
same spectra but express their features in different directions.
Within each context, conditioning on $O$, the context inputs and test token
are sampled identically and independently, while the teacher vector is
sampled independently from its corresponding distribution:
\[
 \vx_1,\ldots,\vx_P,\vx_{\mathrm{test}}
 \overset{\mathrm{i.i.d.}}{\sim}\mathcal N(0,\Sigma),
 \qquad
 \mBeta\sim\mathcal N(0,\Omega).
\]

Define the effective query--key matrix
\begin{equation}
 B_\ell=\frac1N W_x^\top(W_k^\ell)^\top W_q^\ell W_x,
 \qquad \ell=1,\ldots,L,
 \label{eq:app-effective-matching}
\end{equation}
where $N$ is the hidden-state dimension. Before masking, token $i$
assigns context token $j$ the weight
\[
 \frac{(W_q^\ell W_x\vx_i)^\top(W_k^\ell W_x\vx_j)}{NP}
 =\frac{\vx_j^\top B_\ell\vx_i}{P}.
\]
Separate the context and test label states as
\[
 H_y^\ell=[Z_\ell,h_\ell],\qquad
 Z_\ell\in\mathbb R^{N\times P},\quad h_\ell\in\mathbb R^N.
\]
The linear-attention mask defined in Section~\ref{sec:model-setup} gives
\begin{equation}
 \begin{aligned}
 Z_0&=W_y\mBeta^\top U,\qquad h_0=0,\\
 Z_\ell&=Z_{\ell-1}-\frac1{LP}W_vZ_{\ell-1}U^\top B_\ell U,\\
 h_\ell&=h_{\ell-1}+\frac1{LP}W_vZ_{\ell-1}U^\top B_\ell q.
 \end{aligned}
 \label{eq:app-linear-block-recursion}
\end{equation}
Write $Z_\ell=F_\ell U$, where $F_\ell$ maps inputs to context label states.
Substituting
$UU^\top/P=S$ yields
\begin{equation}
 \begin{aligned}
 F_0&=W_y\mBeta^\top,\qquad F_\ell\in\mathbb R^{N\times d},\\
 F_\ell&=F_{\ell-1}-\frac1L W_vF_{\ell-1}SB_\ell,\\
 h_\ell-h_{\ell-1}&=(F_{\ell-1}-F_\ell)q.
 \end{aligned}
 \label{eq:app-linear-compressed-recursion}
\end{equation}
Therefore, the output is
\begin{equation}
 \widehat y_L=W_o\sum_{\ell=1}^L(F_{\ell-1}-F_\ell)q
              =W_o(F_0-F_L)q.
 \label{eq:app-readout-path-expansion}
\end{equation}
The first two steps show how the polynomial is built:
\[
 \begin{aligned}
 F_1&=W_y\mBeta^\top-\frac1L W_vW_y\mBeta^\top SB_1,\\
 F_2&=W_y\mBeta^\top
 -\frac1L W_vW_y\mBeta^\top(SB_1+SB_2)\\
 &\quad+\frac1{L^2}W_v^2W_y\mBeta^\top SB_1SB_2.
 \end{aligned}
\]
With readout coefficients
\[
 m_r=W_oW_v^rW_y,\qquad r=1,\ldots,L,
\]
substitution into the output gives
\begin{equation}
 \mathcal P_L(S)
 =\sum_{r=1}^L\frac{(-1)^{r-1}m_r}{L^r}
   \sum_{1\le\ell_1<\cdots<\ell_r\le L}
   SB_{\ell_1}\cdots SB_{\ell_r}.
 \label{eq:app-noncommutative-polynomial}
\end{equation}
The prediction and easy-task target are
\begin{equation}
 \widehat y_L=\mBeta^\top\mathcal P_L(S)q,
 \qquad y^*=\mBeta^\top q.
 \label{eq:app-general-polynomial-output}
\end{equation}
Thus learning the easy task amounts to approximating the identity on
the relevant input directions. Define
\[
 \mathcal D_L(S)=I_d-\mathcal P_L(S).
\]
Averaging over the teacher at fixed inputs gives
\[
 \begin{aligned}
 \mathbb E_{\mBeta}[(\widehat y_L-y^*)^2\mid U,q,O]
 &=\mathbb E_{\mBeta}[(\mBeta^\top\mathcal D_L(S)q)^2\mid U,q,O]\\
 &=q^\top\mathcal D_L(S)^\top
      \mathbb E[\mBeta\mBeta^\top\mid U,q,O]\mathcal D_L(S)q\\
 &=q^\top\mathcal D_L(S)^\top\Omega\mathcal D_L(S)q.
 \end{aligned}
\]
The evaluation loss $\mathcal R_L$ defined in
\eqref{eq:final-loss} is therefore
\begin{equation}
 \begin{aligned}
 \mathcal R_L
 &=\mathbb E[(\widehat y_L-y^*)^2]\\
 &=\mathbb E_{U,q,O}
    [q^\top\mathcal D_L(S)^\top\Omega\mathcal D_L(S)q].
 \end{aligned}
 \label{eq:app-general-polynomial-risk}
\end{equation}
\subsection{Constrained and Free Polynomial Depth Scaling}
\label{app:polynomial-scaling}

\paragraph{Repeated updates.}
Following the repeated-update construction of \citet{bordelon2026theory}, we
impose isotropic matching and align the shared value map with the label
embedding:
\begin{equation}
 B_\ell=\kappa I_d\quad(\ell=1,\ldots,L),
 \qquad W_vW_y=vW_y.
 \label{eq:app-repeated-update-assumptions}
\end{equation}
Each degree-$r$ product in \eqref{eq:app-noncommutative-polynomial} then equals
$\kappa^rS^r$, and there are $\binom Lr$ choices of $r$ layers. Hence
\begin{equation}
 \begin{gathered}
 \mathcal P_L(S)
 =\sum_{r=1}^L\frac{(-1)^{r-1}m_r}{L^r}
           \binom Lr\kappa^r S^r=p_L(S),\\
 p_L(\lambda)=\sum_{r=1}^L a_r\lambda^r,\qquad
 a_r=(-1)^{r-1}\binom Lr(\kappa/L)^r m_r.
 \end{gathered}
 \label{eq:app-scalar-polynomial}
\end{equation}
In the infinite-context limit $P\to\infty$, we use $S\approx\Sigma$. Thus
\[
 I_d-p_L(\Sigma)
 =O\operatorname{diag}\bigl(1-p_L(\lambda_1),\ldots,
                            1-p_L(\lambda_d)\bigr)O^\top.
\]
Using $\mathbb E[qq^\top\mid O]=\Sigma$ in
\eqref{eq:app-general-polynomial-risk} gives the population-covariance
approximation $\mathcal R_L^{\mathrm{pop}}$ to $\mathcal R_L$:
\begin{equation}
 \begin{aligned}
 \mathcal R_L^{\mathrm{pop}}
 &=\mathbb E_O\operatorname{tr}\!\left[
   \mathcal D_L(\Sigma)^\top\Omega\mathcal D_L(\Sigma)\Sigma\right]\\
 &=\mathbb E_O\operatorname{tr}\!\left[
   O\operatorname{diag}\!\left(
   \lambda_k\omega_k[1-p_L(\lambda_k)]^2
   \right)_{k=1}^dO^\top\right]\\
 &=\sum_{k=1}^d\lambda_k\omega_k[1-p_L(\lambda_k)]^2.
 \end{aligned}
 \label{eq:app-easy-spectral-risk}
\end{equation}
Define $r_L(\lambda)=1-p_L(\lambda)$. Since $p_L(0)=0$, we have
$r_L(0)=1$, and therefore
\begin{equation}
 \deg r_L\le L,\qquad r_L(0)=1,
 \qquad
 \mathcal R_L^{\mathrm{pop}}
 =\sum_{k=1}^d\lambda_k\omega_k r_L(\lambda_k)^2.
 \label{eq:app-identity-target}
\end{equation}

For a specific data spectrum, write
\[
 \begin{gathered}
 \lambda_k=c_\lambda k^{-a},\qquad \omega_k=c_\omega k^{-b},\\
 a\ge0,\qquad b\ge0,
 \qquad \zeta=\frac{a+b-1}{a}\quad(a>0).
 \end{gathered}
\]
where $c_\lambda,c_\omega>0$ and $\lambda_1=c_\lambda$.
For $a>0$, the change of variables from feature index to eigenvalue is
\[
 k=(c_\lambda/\lambda)^{1/a},\qquad
 \left|\frac{dk}{d\lambda}\right|
 =\frac{c_\lambda^{1/a}}{a}\lambda^{-1-1/a}.
\]
Substituting into \eqref{eq:app-identity-target} and replacing the sum by
its continuum integral gives
\begin{equation}
 \begin{aligned}
 \mathcal R_L^{\mathrm{pop}}
 &=c_\lambda c_\omega\sum_{k=1}^d
       k^{-(a+b)}r_L(c_\lambda k^{-a})^2\\
 &\approx c_\lambda c_\omega\int_1^d
       k^{-(a+b)}r_L(c_\lambda k^{-a})^2\,dk\\
 &=\frac{c_\omega c_\lambda^{1-\zeta}}{a}
       \int_{\lambda_d}^{\lambda_1}
       \lambda^{\zeta-1}r_L(\lambda)^2\,d\lambda.
 \end{aligned}
 \label{eq:app-spectral-integral}
\end{equation}
For $a+b>1$, we have $\zeta>0$, which makes the continuum weight integrable
at zero. When $d$ is sufficiently large, $\lambda_d$ is close to zero and the
finite-dimensional cutoff does not yet affect the loss. We may then approximate
the dense spectrum by a measure on $[0,\lambda_1]$. This gives
\begin{equation}
 \mathcal R_L^{\mathrm{pop}}\propto
 \int_0^{\lambda_1}\lambda^{\zeta-1}r_L(\lambda)^2\,d\lambda.
 \label{eq:app-continuum-risk}
\end{equation}

Because $W_v$ is shared across layers,
\[
 m_r=W_oW_v^rW_y\propto v^r,
 \qquad \gamma=\kappa v.
\]
Optimizing $W_oW_y$ to $1$ allows the repeated-update construction to
solve the easy task in the large-depth limit. Substituting into
\eqref{eq:app-scalar-polynomial} gives
\begin{equation}
 \begin{gathered}
 p_L(\lambda)
 =\sum_{r=1}^L(-1)^{r-1}\binom Lr(\gamma\lambda/L)^r
   =1-(1-\gamma\lambda/L)^L,\\
 r_L(\lambda)=(1-\gamma\lambda/L)^L.
 \end{gathered}
 \label{eq:app-constrained-filter}
\end{equation}
For the residual to vanish as $L\to\infty$, the scalar updates must
contract rather than amplify positive-eigenvalue directions. At the
largest eigenvalue this requires
\[
 |1-\gamma\lambda_1/L|<1
 \quad\Longleftrightarrow\quad 0<\gamma<\frac{2L}{\lambda_1}.
\]
Using the optimal-scale estimate of \citet{bordelon2026theory},
\[
 \gamma^*\approx\frac{L}{\lambda_1},\qquad
 \frac{\gamma^*}{L}\approx\frac1{\lambda_1},
\]
the continuum loss becomes, with $u=2L\lambda/\lambda_1$,
\begin{equation}
 \begin{aligned}
 \mathcal R_L^{\mathrm{con}}
 &\approx\frac{c_\omega c_\lambda^{1-\zeta}}{a}
       \int_0^{\lambda_1}\lambda^{\zeta-1}
       (1-\lambda/\lambda_1)^{2L}\,d\lambda\\
 &\sim\frac{c_\omega c_\lambda^{1-\zeta}}{a}
       \left(\frac{\lambda_1}{2L}\right)^\zeta
       \int_0^\infty u^{\zeta-1}e^{-u}\,du\\
 &\propto L^{-\zeta}.
 \end{aligned}
 \label{eq:app-constrained-exponent}
\end{equation}
Here, $\mathcal R_L^{\mathrm{con}}$ denotes the loss of the repeated-update
reference within the population approximation. This exponent is consistent with the
result of \citet{bordelon2026theory}.

\paragraph{Free polynomials.}
We then derive the depth-scaling exponent when the polynomial coefficients are
free by removing the repeated-update conditions in
\eqref{eq:app-repeated-update-assumptions}. We assume only that the depth-$L$
predictor is an arbitrary polynomial in $\Sigma$ of degree at most $L$.
Substitution into
\eqref{eq:app-general-polynomial-risk} gives the freely optimized
population-polynomial reference $\mathcal R_L^{\mathrm{free}}$:
\begin{equation}
 \begin{gathered}
 \mathcal P_L(\Sigma)=p_L(\Sigma),\qquad
 \deg p_L\le L,\qquad p_L(0)=0,
 \qquad S\approx\Sigma,\\
 \mathcal R_L^{\mathrm{pop}}
 =\sum_{k=1}^d\lambda_k\omega_k
   [1-p_L(\lambda_k)]^2,\\
 \mathcal R_L^{\mathrm{free}}\propto
 \min_{\substack{\deg r\le L\\r(0)=1}}
 \int_0^{\lambda_1}\lambda^{\zeta-1}r(\lambda)^2\,d\lambda,
 \qquad r=1-p_L.
 \end{gathered}
 \label{eq:app-free-variational-problem}
\end{equation}
Let $\phi_n$ be a degree-$n$ polynomial orthonormal under this weight:
\[
 \int_0^{\lambda_1}\lambda^{\zeta-1}
       \phi_n(\lambda)\phi_m(\lambda)\,d\lambda=\delta_{nm},
\]
where $\delta_{nm}$ equals one for $n=m$ and zero otherwise. Expanding
the residual and using $r(0)=1$ gives
\[
 \begin{gathered}
 r(\lambda)=\sum_{n=0}^L d_n\phi_n(\lambda),\qquad
 \sum_{n=0}^L d_n\phi_n(0)=1,\\
 \int_0^{\lambda_1}\lambda^{\zeta-1}r(\lambda)^2\,d\lambda
 =\sum_{n=0}^L d_n^2.
 \end{gathered}
\]
Cauchy--Schwarz now gives the minimum directly:
\begin{equation}
 \begin{aligned}
 1&=\left(\sum_{n=0}^L d_n\phi_n(0)\right)^2\\
 &\le\left(\sum_{n=0}^L d_n^2\right)
       \left(\sum_{n=0}^L\phi_n(0)^2\right).
 \end{aligned}
\end{equation}
Equality is attained by coefficients proportional to the endpoint values:
\begin{equation}
 d_n^*=\frac{\phi_n(0)}{\sum_{m=0}^L\phi_m(0)^2},\qquad
 \mathcal R_L^{\mathrm{free}}\propto
       \left[\sum_{n=0}^L\phi_n(0)^2\right]^{-1}.
 \label{eq:app-polynomial-kernel-solution}
\end{equation}
For this weight, the basis consists of shifted Jacobi polynomials:
\begin{equation}
 \phi_n(\lambda)=\sqrt{\frac{2n+\zeta}{\lambda_1^\zeta}}\,
       P_n^{(\zeta-1,0)}(1-2\lambda/\lambda_1),
 \label{eq:app-shifted-jacobi-basis}
\end{equation}
where $P_n^{(\zeta-1,0)}$ is the degree-$n$ Jacobi polynomial.
Using its value at $1$ gives
\[
 \begin{aligned}
 \phi_n(0)^2
 &=\frac{2n+\zeta}{\lambda_1^\zeta}
   \left[\frac{\Gamma(n+\zeta)}
   {\Gamma(\zeta)\Gamma(n+1)}\right]^2\\
 &\sim\frac{2}{\lambda_1^\zeta\Gamma(\zeta)^2}\,n^{2\zeta-1}.
 \end{aligned}
\]
Consequently,
\begin{equation}
 \begin{aligned}
 \sum_{n=0}^L\phi_n(0)^2
 &\sim\frac{2}{\lambda_1^\zeta\Gamma(\zeta)^2}
       \int_0^L n^{2\zeta-1}\,dn\\
 &=\frac{L^{2\zeta}}{\zeta\lambda_1^\zeta\Gamma(\zeta)^2}.
 \end{aligned}
 \label{eq:app-free-exact-risk}
\end{equation}
The two polynomial families therefore give
\begin{equation}
 \mathcal R_L^{\mathrm{con}}\propto L^{-\zeta},\qquad
 \mathcal R_L^{\mathrm{free}}\propto L^{-2\zeta},\qquad
 \zeta=\frac{a+b-1}{a}.
 \label{eq:app-double-exponent}
\end{equation}

Table~\ref{tab:polynomial-exponent-comparison} compares these reference
exponents with the measured values.

\begin{table}[t]
 \centering
 \caption{Measured depth-scaling exponents of the linear-attention model on
 the easy task lie between two reference values.}
 \label{tab:polynomial-exponent-comparison}
 \vspace{\baselineskip}
 \begin{tabular}{cccc}
 \toprule
 $(a,b)$ & Constrained $\zeta$ & Measured $\alpha$ & Free $2\zeta$\\
 \midrule
 $(2,0.5)$ & $0.750$ & $1.121$ & $1.500$\\
 $(2,1)$   & $1.000$ & $1.485$ & $2.000$\\
 $(2,1.5)$ & $1.250$ & $1.684$ & $2.500$\\
 $(2,2)$   & $1.500$ & $1.987$ & $3.000$\\
 $(0.5,2)$ & $3.000$ & $3.310$ & $6.000$\\
 $(1,2)$   & $2.000$ & $2.552$ & $4.000$\\
 $(1.5,2)$ & $1.667$ & $2.163$ & $3.333$\\
 \bottomrule
 \end{tabular}
\end{table}

The actual model is less constrained than the repeated-update construction
of \citet{bordelon2026theory}, which relies on the two conditions in
\eqref{eq:app-repeated-update-assumptions}, but it is not as unconstrained as the idealized
free polynomial, whose coefficients are optimized independently. It is
therefore reasonable for the measured exponents to lie between these two
reference values.

Note that this theoretical analysis has several requirements, including the
population-covariance approximation $S\approx\Sigma$, corresponding to
$P\to\infty$; a dense spectrum with $\zeta>0$; and a sufficiently large
spectral range $\lambda_1/\lambda_d$. At the evaluated depths, repeated
updates require $1\ll L\ll\lambda_1/\lambda_d$, whereas free polynomials
require $1\ll L\ll\sqrt{\lambda_1/\lambda_d}$.
\subsection{Predicting the Linear-Model Loss Floor}
\label{app:difficult-linear-geometry}

In this subsection, we characterize the predictable component of the
difficult-task target and the resulting irreducible loss floor for the
linear-attention model. We further validate these theoretical predictions
using Monte Carlo estimates.

For a test token $q$ sampled independently of the context inputs, the
difficult-task target is
\begin{equation}
 \begin{gathered}
 S_k(q)=\operatorname{TopK}_{1\le i\le P}
      \frac{q^\top\vx_i}{\|q\|\,\|\vx_i\|},\\
 t=\frac1k\sum_{i\in S_k(q)}\vx_i\in\mathbb R^d,
 \qquad y^*=\mBeta^\top t.
 \end{gathered}
 \label{eq:app-difficult-target-vector}
\end{equation}
From Appendix~\ref{app:linear-polynomial-theory}, assuming $S\approx\Sigma$,
\begin{equation}
 \begin{aligned}
 \widehat y_L=\mBeta^\top p_L(\Sigma)q,
 \qquad p_L(\Sigma)=\sum_{r=1}^L a_r\Sigma^r.
 \end{aligned}
 \label{eq:app-difficult-polynomial-output}
\end{equation}
Write $u_L(q)=p_L(\Sigma)q$. Since
$\widehat y_L-y^*=\mBeta^\top(u_L-t)$, averaging first over the
independent teacher gives
\[
 \begin{aligned}
 \mathbb E_{\mBeta}[(\widehat y_L-y^*)^2\mid U,q,O]
 &=(t-u_L)^\top\mathbb E[\mBeta\mBeta^\top\mid O](t-u_L)\\
 &=(t-u_L)^\top\Omega(t-u_L).
 \end{aligned}
\]
Thus $\mathcal R_L^{\mathrm{pop}}=\mathbb E\|t-u_L\|_\Omega^2$,
where $\|z\|_\Omega^2=z^\top\Omega z$.
At fixed $q,O$, the selected target $t$ varies across contexts, whereas
$u_L$ does not. Separate its conditional mean from this variation:
\[
 \begin{gathered}
 \bar t=\mathbb E[t\mid q,O],\qquad \delta=t-\bar t,\\
 \mathbb E[\delta\mid q,O]=0.
 \end{gathered}
\]
Substituting $t-u_L=\delta+(\bar t-u_L)$ and expanding gives
\[
 \|t-u_L\|_\Omega^2
 =\|\delta\|_\Omega^2+\|\bar t-u_L\|_\Omega^2
   +2\delta^\top\Omega(\bar t-u_L).
\]
The cross term vanishes after conditioning because $\Omega$ and
$\bar t-u_L$ are fixed given $q,O$:
\[
 \begin{aligned}
 \mathbb E\!\left[\delta^\top\Omega
       \bigl(\bar t-p_L(\Sigma)q\bigr)\right]
 &=\mathbb E\!\left[\mathbb E[\delta\mid q,O]^\top\Omega
       \bigl(\bar t-p_L(\Sigma)q\bigr)\right]\\
 &=0.
 \end{aligned}
\]
Hence the population risk separates into
\begin{equation}
 \begin{aligned}
 \mathcal R_L^{\mathrm{pop}}
 &=\mathbb E\|t-p_L(\Sigma)q\|_\Omega^2\\
 &=\underbrace{\mathbb E\|\delta\|_\Omega^2}
       _{\text{context-specific variation}}
   +\underbrace{\mathbb E\|\bar t-p_L(\Sigma)q\|_\Omega^2}
       _{\text{polynomial approximation error}}.
 \end{aligned}
 \label{eq:app-difficult-weighted-risk}
\end{equation}
We denote the first term by $\mathcal R_{\mathrm{irr}}$. It is independent of
the polynomial degree, whereas increasing the degree can reduce only the
second term.

\paragraph{Predictable component of the difficult-task target.}
Section~\ref{sec:difficult-linear} states that the predictable component is
approximately proportional to $\Sigma q$. We now derive this result by
approximating cosine-similarity ranking with the inner-product scores
$s_i=q^\top\vx_i$. Conditional on $q,O$, the pair $(\vx_i,s_i)$ is jointly
Gaussian, with
\[
 \mathbb E[s_i\mid q,O]=0,\qquad
 \operatorname{Var}(s_i\mid q,O)=q^\top\Sigma q=:V(q),\qquad
 \operatorname{Cov}(\vx_i,s_i\mid q,O)=\Sigma q.
\]
The Gaussian conditioning formula therefore gives
\begin{equation}
 \mathbb E[\vx_i\mid s_i,q,O]
 =\frac{\operatorname{Cov}(\vx_i,s_i\mid q,O)}
        {\operatorname{Var}(s_i\mid q,O)}s_i
 =\frac{\Sigma q}{V(q)}s_i.
 \label{eq:app-gaussian-score-regression}
\end{equation}
Let $s_{(1)}\ge\cdots\ge s_{(P)}$ be the ordered scores, and let
$\vx_{(j)}$ denote the input attaining $s_{(j)}$. Since
$z_i=s_i/\sqrt{V(q)}$ are independent standard normal variables, define
\[
 \gamma_{P,k}
 =\mathbb E\!\left[\frac1k\sum_{j=1}^k z_{(j)}\right].
\]
Applying \eqref{eq:app-gaussian-score-regression} to the $k$ selected inputs
and averaging over their ordered scores gives
\begin{equation}
 \begin{aligned}
 \bar t(q)
 &=\mathbb E\!\left[\frac1k\sum_{j=1}^k\vx_{(j)}\Bigm|q,O\right]\\
 &=\frac{\Sigma q}{V(q)}
   \mathbb E\!\left[\frac1k\sum_{j=1}^k s_{(j)}\Bigm|q,O\right]
 =\gamma_{P,k}\frac{\Sigma q}{\sqrt{q^\top\Sigma q}}.
 \end{aligned}
 \label{eq:app-inner-product-mean}
\end{equation}
To remove the remaining query-dependent denominator, note that
for $q\sim\mathcal N(0,\Sigma)$,
\[
 \mathbb E[V(q)\mid O]=\operatorname{tr}(\Sigma^2),\qquad
 \operatorname{Var}(V(q)\mid O)=2\operatorname{tr}(\Sigma^4).
\]
When $2\operatorname{tr}(\Sigma^4)/\operatorname{tr}(\Sigma^2)^2\ll1$,
we can replace $V(q)$ by its mean. Combining this with the cosine-to-score
approximation yields
\[
 \bar t(q)\approx c\Sigma q,\qquad
 c=\frac{\gamma_{P,k}}{\sqrt{\operatorname{tr}(\Sigma^2)}}.
\]
Since the teacher is independent of the selected inputs given $q,O$,
\begin{equation}
 \begin{gathered}
 \mathbb E[y^*\mid q,\mBeta,O]
 =\mBeta^\top\mathbb E[t\mid q,O]
 \approx c\,\mBeta^\top\Sigma q.
 \end{gathered}
 \label{eq:app-difficult-leading-target}
\end{equation}
Thus, the predictable component is proportional to $\Sigma q$ and can already
be represented by the degree-one polynomial $p_1(\Sigma)=c\Sigma$. Higher
polynomial degrees cannot remove the context-specific variation in
\eqref{eq:app-difficult-weighted-risk}, explaining both the rapid saturation
and the nonzero loss floor of the linear-attention model.

\paragraph{An explicit expression for the irreducible loss.}
Let
$\mathbf a=(a_1,\ldots,a_L)^\top$ collect the polynomial coefficients and
define
\begin{equation}
 \begin{aligned}
 \Phi_L(q)=[\Sigma q,\ldots,\Sigma^Lq],\qquad
 p_L(\Sigma)q=\Phi_L(q)\mathbf a.
 \end{aligned}
 \label{eq:app-polynomial-features}
\end{equation}
The corresponding population loss expands as
\begin{equation}
 \begin{aligned}
 \mathcal R_L^{\mathrm{pop}}(\mathbf a)
 &=\mathbb E\|t-\Phi_L(q)\mathbf a\|_\Omega^2\\
 &=\mathbb E[t^\top\Omega t]
   -2\mathbf a^\top\mathbb E[\Phi_L(q)^\top\Omega t]\\
 &\quad+\mathbf a^\top
   \mathbb E[\Phi_L(q)^\top\Omega\Phi_L(q)]\mathbf a\\
 &=T-2\mathbf a^\top h+\mathbf a^\top G\mathbf a,
 \end{aligned}
 \label{eq:app-polynomial-quadratic}
\end{equation}
where
\begin{equation}
 T=\mathbb E[t^\top\Omega t],\qquad
 h=\mathbb E[\Phi_L(q)^\top\Omega t],\qquad
 G=\mathbb E[\Phi_L(q)^\top\Omega\Phi_L(q)].
 \label{eq:app-polynomial-floor-statistics}
\end{equation}
In particular, the entries of the Gram matrix are determined by the spectra:
\begin{equation}
 G_{rs}
 =\mathbb E[(\Sigma^rq)^\top\Omega(\Sigma^sq)]
 =\operatorname{tr}(\Omega\Sigma^{r+s+1}),
 \qquad 1\le r,s\le L.
 \label{eq:app-polynomial-gram}
\end{equation}
Differentiating \eqref{eq:app-polynomial-quadratic} with respect to
$\mathbf a$ gives the normal equations
$G\mathbf a=h$. Hence
\begin{equation}
 \mathbf a_*=G^\dagger h,\qquad
 \mathcal R_{L,*}^{\mathrm{pop}}=T-h^\top G^\dagger h.
 \label{eq:app-polynomial-quadratic-minimum}
\end{equation}
The decomposition in \eqref{eq:app-difficult-weighted-risk} also gives
\begin{equation}
 \min_{\mathbf a}\mathcal R_L^{\mathrm{pop}}(\mathbf a)
 =\mathcal R_{\mathrm{irr}}
 +\min_{\mathbf a}\mathbb E
   \|\bar t-\Phi_L(q)\mathbf a\|_\Omega^2.
 \label{eq:app-floor-vs-irreducible}
\end{equation}
Under the degree-one approximation
$\bar t\approx c\Sigma q$ in
\eqref{eq:app-difficult-leading-target}, the second term is negligible.
Combining \eqref{eq:app-polynomial-quadratic-minimum} and
\eqref{eq:app-floor-vs-irreducible} therefore gives the explicit prediction
\begin{equation}
 \mathcal R_{\mathrm{irr}}
 \approx T-h^\top G^\dagger h.
 \label{eq:app-explicit-irreducible-floor}
\end{equation}

We evaluate this prediction by Monte Carlo sampling in the original cosine
task. We independently sample $n$ query--rotation--context triples. For sample
$i$, $t_i$ is the resulting cosine-selected target and
$\Omega_i=O_i\operatorname{diag}(\omega_1,\ldots,\omega_d)O_i^\top$.
The hats below denote empirical estimates computed from these samples:
\begin{equation}
 \begin{gathered}
 \widehat T=\frac1n\sum_{i=1}^n t_i^\top\Omega_i t_i,\qquad
 \widehat h=\frac1n\sum_{i=1}^n\Phi_L(q_i)^\top\Omega_i t_i,\\
 \widehat G=\frac1n\sum_{i=1}^n
 \Phi_L(q_i)^\top\Omega_i\Phi_L(q_i),\qquad
 \widehat{\mathcal R}_{\mathrm{irr}}
 =\widehat T-\widehat h^\top\widehat G^\dagger\widehat h.
 \end{gathered}
 \label{eq:app-monte-carlo-polynomial-floor}
\end{equation}
Table~\ref{tab:linear-floor-validation} compares this Monte Carlo prediction
of \eqref{eq:app-explicit-irreducible-floor} with the actual loss plateau of
the trained model. Their close agreement validates the predicted irreducible
floor.

\begin{table}[t]
 \centering
 \caption{The theoretically predicted irreducible loss closely matches the
 actual loss plateau of the linear-attention model.}
 \label{tab:linear-floor-validation}
 \vspace{\baselineskip}
 \small
 \begin{tabular}{ccrr}
 \toprule
 $(a,b)$ & Theory-predicted loss & Measured loss & Difference \\
 \midrule
 \multicolumn{4}{l}{$k=2$} \\
 $(0,0)$       & $59.3714\pm0.0213$ & 59.8502 & $0.80\%$ \\
 $(0.25,0.25)$ & $9.7004\pm0.0049$  & 9.7387  & $0.39\%$ \\
 $(0.5,0.5)$   & $2.2715\pm0.0061$  & 2.3274  & $2.40\%$ \\
 $(1,0.25)$    & $1.2246\pm0.0106$  & 1.2259  & $0.11\%$ \\
 $(1,1)$       & $0.4975\pm0.0102$  & 0.5036  & $1.22\%$ \\
 \midrule
 \multicolumn{4}{l}{$k=16$} \\
 $(0.5,0.5)$   & $0.3218\pm0.0042$  & 0.3083  & $4.38\%$ \\
 $(1,0)$       & $0.4481\pm0.0086$  & 0.4452  & $0.67\%$ \\
 $(1,0.25)$    & $0.3186\pm0.0085$  & 0.3222  & $1.11\%$ \\
 $(1.25,0)$    & $0.3595\pm0.0090$  & 0.3659  & $1.75\%$ \\
 \bottomrule
 \end{tabular}
\end{table}

\begin{figure}[t]
 \centering
 \includegraphics[width=0.58\linewidth]{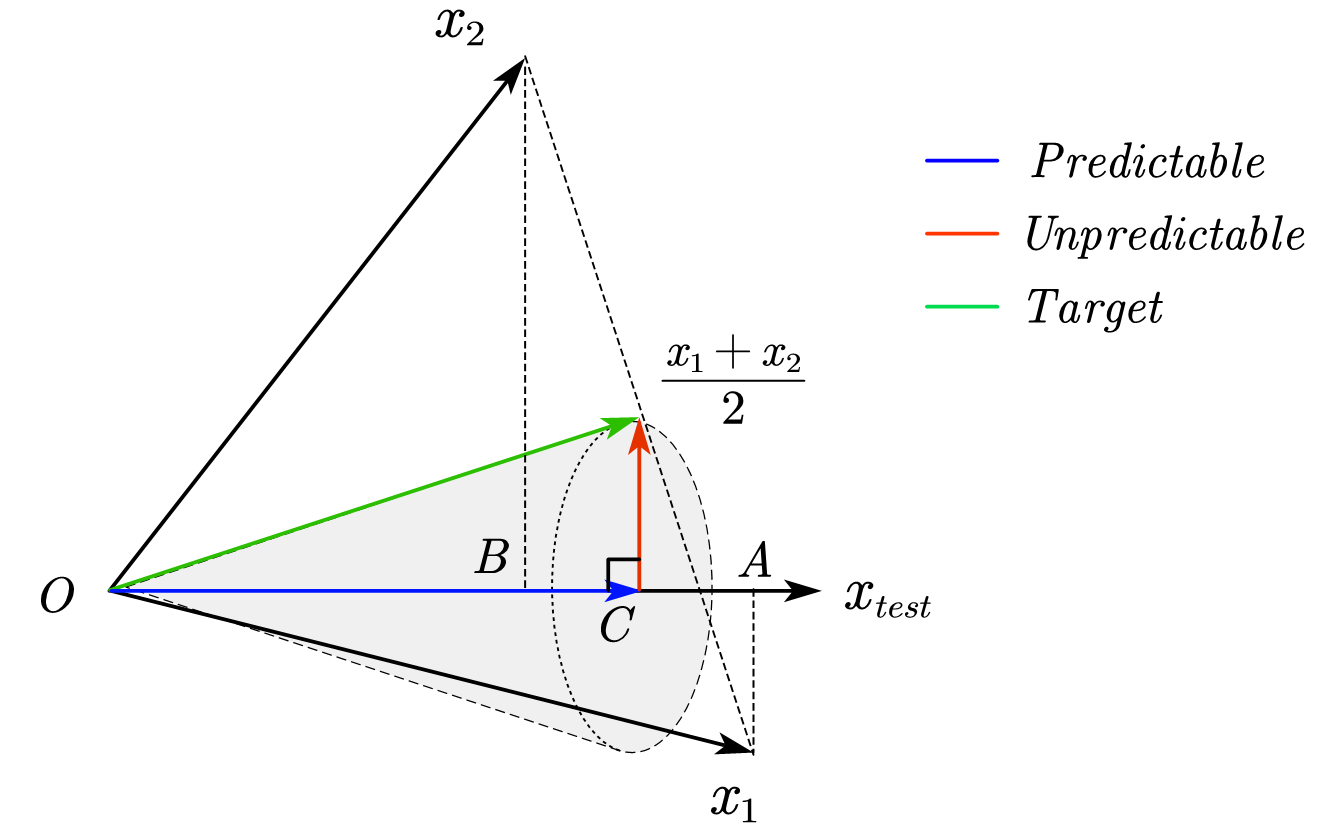}
 \caption{The linear-attention model captures only the average predictable
 component of the difficult-task target. The green target is the average of
 the two selected inputs. Blue shows the best predictable component
 $\overrightarrow{OC}$; red shows the context-specific residual.}
 \label{fig:difficult-linear-geometry}
\end{figure}

\paragraph{Geometric interpretation.}
We now provide a heuristic geometric picture of the predictable component
and irreducible loss, using the top-$2$ task as an example. In
Figure~\ref{fig:difficult-linear-geometry}, the horizontal ray represents the
query $x_{\mathrm{test}}$. The black vectors $\overrightarrow{Ox_1}$ and
$\overrightarrow{Ox_2}$ are the two selected context inputs, and their average
is the green target $\overrightarrow{Ot}=(x_1+x_2)/2$. The blue vector
$\overrightarrow{OC}$ represents the conditional mean $\bar t$, whereas the
red vector $\overrightarrow{Ct}=t-\bar t$ is the context-specific residual.
Linear attention can approximate the predictable blue component through the
population covariance, but it lacks the context-specific selection required
to follow the red residual for each realized context. This residual therefore
produces the irreducible loss floor.

\subsection{Power Laws from Mode-Wise Exponential Decay}
\label{app:exponential-superposition}

As a heuristic model of Figure~\ref{fig:easy-linear-depth-scaling}(c),
we show how exponential decay at different rates can combine into a
power law. Let feature $k$ have initial loss weight $w_k\ge0$ and
characteristic depth $\tau_k>0$:
\[
 \mathcal R_k(L)=w_ke^{-L/\tau_k},\qquad
 \mathcal R_L=\sum_k w_ke^{-L/\tau_k}.
\]
Assume the loss-weighted distribution of characteristic depths follows
a power law, $\rho(\tau)=C\tau^{-\beta}$ for $\tau\ge\tau_0>0$, with
$C>0$ and $\beta>1$. The aggregate loss is
\begin{equation}
 \mathcal R_L\approx\int_{\tau_0}^\infty
       \rho(\tau)e^{-L/\tau}\,d\tau.
 \label{eq:exponential-mode-mixture}
\end{equation}
With $x=L/\tau$ and $d\tau=-Lx^{-2}\,dx$, this becomes
\begin{equation}
 \begin{aligned}
 \mathcal R_L&\approx CL^{1-\beta}
    \int_0^{L/\tau_0}x^{\beta-2}e^{-x}\,dx\\
 &\sim C\Gamma(\beta-1)L^{-(\beta-1)},\qquad L\gg\tau_0.
 \end{aligned}
 \label{eq:exponential-mixture-power-law}
\end{equation}
Thus exponential convergence of individual features produces power-law
depth scaling when their characteristic depths have a broad power-law
distribution. Stronger features converge first, while weaker features
continue to contribute as depth increases.

\section{Nonlinear-Attention Model Theory}
\label{app:theory}

In this section, we derive the loss decomposition for the
nonlinear-attention model on the difficult task studied in
Section~\ref{sec:difficult-nonlinear}. 

We first introduce the notation. Throughout this section, $\mathbb E$ denotes
an expectation over contexts. Let $L$ be the number of layers and $P$ the
number of context tokens.
We use $\ell=1,\ldots,L$ for layers and $i=1,\ldots,P$ for tokens.
Write $k$ for the number of labels selected by the
top-$k$ task. We measure the relevance of each context token to the test
token using cosine similarity:
\begin{equation}
 \begin{gathered}
 s_i=\frac{\vx_{\mathrm{test}}^\top\vx_i}
              {\|\vx_{\mathrm{test}}\|_2\|\vx_i\|_2},\\
 s_{i_{(1)}}\ge\cdots\ge s_{i_{(P)}},
 \qquad S_k=\{i_{(1)},\ldots,i_{(k)}\}.
 \end{gathered}
 \label{eq:app-target-selection}
\end{equation}
Thus $S_k$ contains the $k$ context indices closest to the test token.
For the label column $y=(y_1,\ldots,y_P)^\top$, define the target row
$t_k=(t_{k,1},\ldots,t_{k,P})$ by
\begin{equation}
 \begin{gathered}
 t_{k,i}=
 \begin{cases}
  1/k,&i\in S_k,\\
  0,&i\notin S_k,
 \end{cases}\\
 y^*=t_ky=\frac1k\sum_{i\in S_k}y_i .
 \end{gathered}
 \label{eq:app-target-weights}
\end{equation}

Let $p_{\ell i}$ be the attention weight from the test token to context
token $i$ at layer $\ell$, with softmax normalized over the $P$ context sources
since the test token does not attend to itself:
\begin{equation}
 p_\ell=(p_{\ell1},\ldots,p_{\ell P}),
 \qquad \sum_{i=1}^P p_{\ell i}=1.
\end{equation}
With fixed context states, each layer reads the same label column $y$.
The common scalar gain $g_L$ includes the value map and readout:
\begin{equation}
 \begin{gathered}
 \widehat y_L=\frac1L\sum_{\ell=1}^Lf_\ell,\\
 f_\ell=g_Lp_\ell y=g_L\sum_{i=1}^Pp_{\ell i}y_i,\\
 g_L=W_oW_vW_y.
 \end{gathered}
 \label{eq:app-layer-predictor}
\end{equation}
Define the layer error relative to the same target as $e_\ell=f_\ell-y^*$.
For $L>1$, assuming finite expected squared errors, let
\begin{equation}
 \begin{gathered}
 A_L=\frac1L\sum_{\ell=1}^L\mathbb E[e_\ell^2],\\
 C_L=\frac1{L(L-1)}\sum_{\ell\ne m}\mathbb E[e_\ell e_m].
 \end{gathered}
 \label{eq:layer-error-moments}
\end{equation}
The second sum is over ordered distinct pairs. Since
$\widehat y_L-y^*=L^{-1}\sum_\ell e_\ell$, expanding the square gives
\begin{equation}
 \begin{aligned}
 \mathcal R_L
 &=\mathbb E[(\widehat y_L-y^*)^2]\\
 &=\frac1{L^2}\left[
       \sum_\ell\mathbb E[e_\ell^2]
       +\sum_{\ell\ne m}\mathbb E[e_\ell e_m]\right]\\
 &=\frac{A_L}{L}+\frac{L-1}{L}C_L\\
 &=C_L+\frac{A_L-C_L}{L}.
 \end{aligned}
 \label{eq:appendix-layerwise-risk}
\end{equation}
Expanding pairwise differences and using \eqref{eq:layer-error-moments},
\[
 \begin{aligned}
 \sum_{\ell\ne m}\mathbb E[(e_\ell-e_m)^2]
 &=2(L-1)\sum_\ell\mathbb E[e_\ell^2]
   -2\sum_{\ell\ne m}\mathbb E[e_\ell e_m]\\
 &=2L(L-1)(A_L-C_L).
 \end{aligned}
\]
Dividing by $2L(L-1)$ gives
\begin{equation}
 A_L-C_L
 =\frac1{2L(L-1)}
       \sum_{\ell\ne m}\mathbb E[(e_\ell-e_m)^2]\ge0 .
 \label{eq:pairwise-layer-disagreement}
\end{equation}

$C_L$ captures the error shared across layers, whereas $A_L-C_L$ measures
their disagreement. Averaging reduces the latter contribution to
$(A_L-C_L)/L$. The shared component cannot be removed through aggregation
and therefore gives rise to the irreducible loss, while differences among
layer errors can partially cancel and produce the reducible loss.

Importantly, this decomposition does not by itself imply inverse-depth scaling. We first establish the emergence of inverse-depth scaling empirically through systematic experiments and rigorous power-law fitting, and then use the decomposition to identify the components and mechanisms underlying this behavior.

This decomposition relies on the parallel interpretation of our controlled
model: the layers act as independently parameterized predictors whose
aggregate forms the final output. A layerwise decomposition is therefore
natural in this setting. In the generalized settings considered below,
however, it is neither the only possible decomposition nor necessarily the
most natural one. For example, when inputs and labels share a single stream,
the model becomes cascaded and each layer depends on the preceding layers.
A mean-field treatment \citep{mei2018meanfield,chizat2018global,rotskoff2018parameters}
may then be more appropriate, separating each layer's output into a mean
component coupled to the other layers and a layer-specific fluctuation.
Developing a rigorous decomposition for
more complex systems, including large language models, remains an important
direction for understanding the role of depth in realistic models.

\section{Generalized Model Experiments and Analysis}
\label{app:generalized-models}

We now relax some design choices from the controlled model in
Section~\ref{sec:model-setup} to move the architecture closer to standard
Transformers and LLMs. First, the original two-stream encoding separates token
matching from label propagation: queries and keys interact only with the
input stream, whereas values and the output map interact only with the label
stream. Second, all layers share a value matrix, which improves training
stability and keeps the model concise. Third, the nonlinear-attention model
initially keeps the context states fixed, isolating the process by which
different layers retrieve relevant information from the context. In the
following subsections, we relax these assumptions one at a time and conduct
analogous experiments and analyses for each extension.

\subsection{Encoding Input and Label Information into One Space}
\label{app:single-stream}

Section~\ref{sec:model-setup} initializes separate input and label streams,
$H_x=W_xX$ and $H_y^0=W_y\vy^{\mathrm{in}}$. In standard LLMs, token and
label information are commonly encoded in a shared representation space. We
therefore embed both sources of information in a shared token state:
\begin{equation}
  H^0=W_xX+W_y\vy^{\mathrm{in}}
  \in\mathbb R^{N\times(P+1)}.
  \label{eq:single-stream-encoding}
\end{equation}
Queries, keys, and values are now all computed from the same evolving state:
\begin{equation}
  Q^\ell=W_q^\ell H^{\ell-1},
  \qquad
  K^\ell=W_k^\ell H^{\ell-1},
  \qquad
  V^\ell=W_vH^{\ell-1}.
  \label{eq:single-stream-qkv}
\end{equation}
For nonlinear attention, the shared state is updated as
\begin{equation}
  \begin{aligned}
    A^\ell
    &=\mathsf M\odot\operatorname{softmax}\!\left((Q^\ell)^\top K^\ell\right),\\
    H^\ell
    &=H^{\ell-1}+\frac{1}{L}V^\ell(A^\ell)^\top.
  \end{aligned}
  \label{eq:single-stream-attention}
\end{equation}
Thus token matching, information propagation, and residual updates all act
within a common representation space, and each layer operates on the state
produced by the preceding layers.

To define the two decomposition terms in this cascaded model, let
$z_\ell=W_oh_{\mathrm{test}}^\ell$ denote the test-token readout after
layer $\ell$. Define the base readout and layer increments as
\begin{equation}
  b_L=z_0,
  \qquad
  d_\ell=z_\ell-z_{\ell-1},
  \qquad
  \widehat y_L=b_L+\sum_{\ell=1}^Ld_\ell.
  \label{eq:single-stream-increments}
\end{equation}
We can express the final prediction as an average of effective layerwise
predictors by setting
\begin{equation}
  f_\ell=b_L+Ld_\ell,
  \qquad
  e_\ell=f_\ell-y^*,
  \qquad
  \widehat y_L-y^*=\frac1L\sum_{\ell=1}^Le_\ell.
  \label{eq:single-stream-effective-predictors}
\end{equation}
For $L>1$, the shared-error and disagreement terms are then defined as
\begin{equation}
  \begin{aligned}
    C_L
    &=\frac{1}{L(L-1)}
      \sum_{\ell\ne m}\mathbb E[e_\ell e_m],\\
    A_L-C_L
    &=\frac{1}{2L(L-1)}
      \sum_{\ell\ne m}\mathbb E[(e_\ell-e_m)^2]\\
    &=\frac{1}{L-1}\sum_{\ell=1}^L
      \mathbb E[(f_\ell-\widehat y_L)^2].
  \end{aligned}
  \label{eq:single-stream-error-components}
\end{equation}
With these definitions, the loss retains the exact identity
\begin{equation}
  \mathcal R_L=C_L+\frac{A_L-C_L}{L}.
  \label{eq:single-stream-loss-decomposition}
\end{equation}

\begin{figure}[t]
  \centering
  \includegraphics[width=\linewidth]{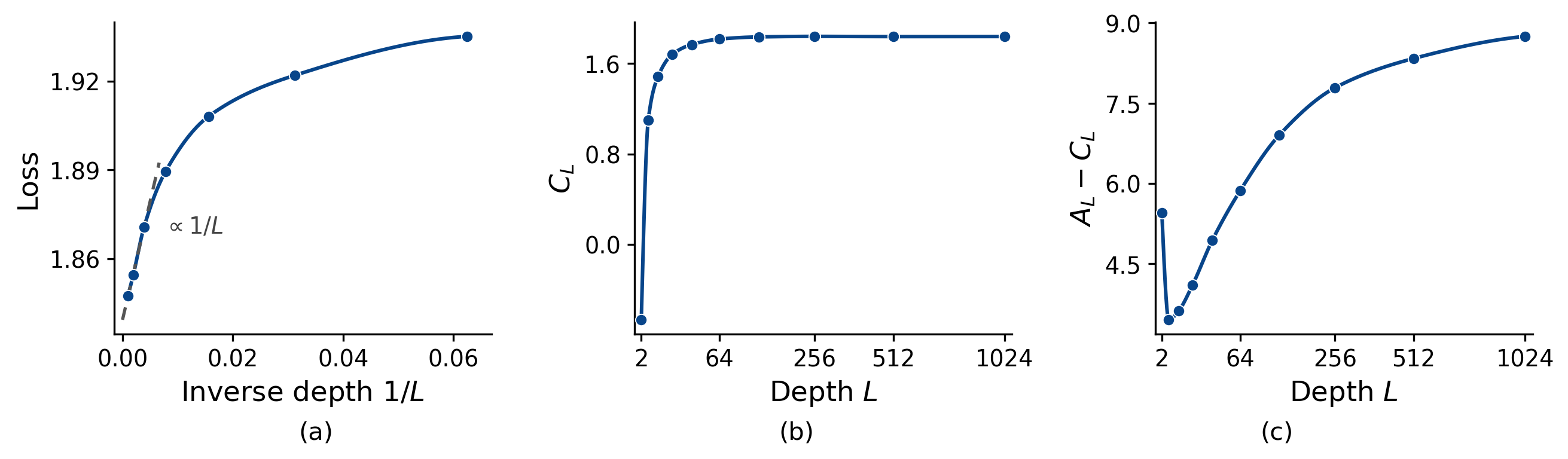}
  \caption{The single-stream nonlinear-attention model exhibits a similar
  inverse-depth scaling signal. (a) The model is evaluated under isotropic data spectra. (b,c) The
  shared-error term $C_L$ and disagreement term $A_L-C_L$ both approach finite
  constants with depth.}
  \label{fig:single-stream-diagnostics}
\end{figure}

We evaluate this generalized model on the difficult task.
Figure~\ref{fig:single-stream-diagnostics} shows that $C_L$ and $A_L-C_L$
continue to approach finite constants, while the loss exhibits an approximate
inverse-depth trend beyond moderate depth.

\subsection{Layer-Decoupled Value Matrices}
\label{app:untied-values}

In standard LLMs, value matrices are typically parameterized independently
across layers. We therefore replace the value matrix $W_v$ shared in
Section~\ref{sec:model-setup} with independently parameterized matrices
$W_v^\ell$, giving
\begin{equation}
  Q^\ell=W_q^\ell H_x,
  \qquad
  K^\ell=W_k^\ell H_x,
  \qquad
  V^\ell=W_v^\ell H_y^{\ell-1}.
  \label{eq:untied-value-qkv}
\end{equation}
This modification allows each layer to apply a distinct transformation to
the retrieved label information.

\begin{figure}[t]
  \centering
  \includegraphics[width=\linewidth]{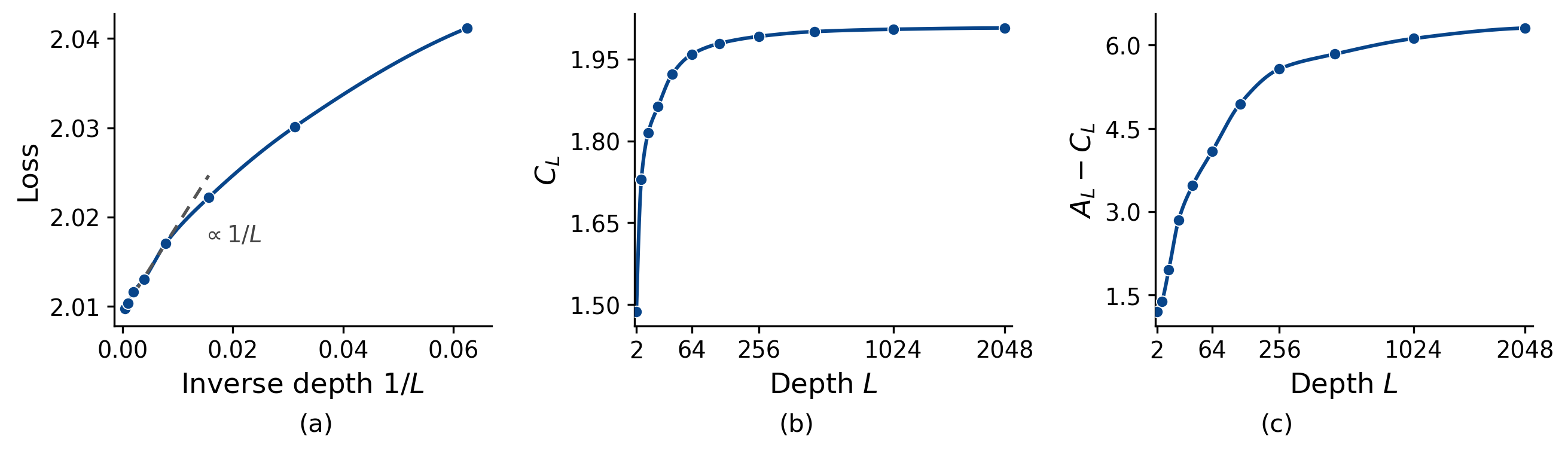}
  \caption{The nonlinear-attention model with layer-specific value matrices
  exhibits a similar inverse-depth scaling signal. (a) The model is evaluated
  under isotropic data spectra.
  (b,c) The shared-error term $C_L$ and disagreement term $A_L-C_L$ both
  approach finite constants with depth.}
  \label{fig:untied-value-diagnostics}
\end{figure}

Let $p_\ell$ denote the attention weights from the test token to the context
tokens, and let $y$ collect their labels. The corresponding layerwise
predictor and error are
\begin{equation}
  g_\ell=W_oW_v^\ell W_y,
  \qquad
  f_\ell=g_\ell p_\ell y,
  \qquad
  e_\ell=f_\ell-y^*,
  \qquad
  \widehat y_L=\frac{1}{L}\sum_{\ell=1}^Lf_\ell.
  \label{eq:untied-value-predictors}
\end{equation}
For $L>1$, the shared-error and disagreement terms are
\begin{equation}
  \begin{aligned}
    C_L
    &=\frac{1}{L(L-1)}
      \sum_{\ell\ne m}\mathbb E[e_\ell e_m],\\
    A_L-C_L
    &=\frac{1}{2L(L-1)}
      \sum_{\ell\ne m}\mathbb E[(e_\ell-e_m)^2]\\
    &=\frac{1}{2L(L-1)}
      \sum_{\ell\ne m}
      \mathbb E\!\left[\bigl((g_\ell p_\ell-g_mp_m)y\bigr)^2\right].
  \end{aligned}
  \label{eq:untied-value-error-components}
\end{equation}
The loss therefore retains the exact identity
\begin{equation}
  \mathcal R_L=C_L+\frac{A_L-C_L}{L}.
  \label{eq:untied-value-loss-decomposition}
\end{equation}
We evaluate this generalized model on the difficult task.
Figure~\ref{fig:untied-value-diagnostics} shows that $C_L$ and $A_L-C_L$
continue to approach finite constants, while the loss displays
an approximate inverse-depth trend beyond moderate depth. Thus, the same
inverse-depth scaling signal remains visible after relaxing the shared-value
assumption.

\subsection{Allowing Context Updates in the Nonlinear-Attention Model}
\label{app:context-updates}

Our linear-attention model in Section~\ref{sec:model-setup} follows
\citet{bordelon2026theory}, including their attention-masking scheme.
Let $\bm{1}_P\in\mathbb R^P$ denote the all-ones vector. For a single
test token, their mask can be written as
\begin{equation}
  \mathsf M
  =
  \begin{pmatrix}
    -\bm{1}_P\bm{1}_P^\top & \bm{0}_{P\times 1}\\
    \bm{1}_P^\top & 0
  \end{pmatrix},
  \label{eq:bordelon-context-mask}
\end{equation}
where the upper-left block governs context-to-context updates, while the
lower-left block allows the test token to attend to the context tokens.
Intuitively, this sign structure lets linear attention extract data structure
at progressively finer spectral resolutions. The negative context-to-context
block preferentially subtracts components shared by highly similar context
tokens from their label states. Dominant spectral features produce larger
similarity-weighted updates
and are therefore resolved earlier, whereas weaker features persist until
greater depth and can be learned in deeper layers. This behavior is
consistent with the sequential mode learning in
Figure~\ref{fig:easy-linear-depth-scaling}(c). By contrast, the positive
test-to-context block allows retrieved information to accumulate in the test
token's label state, which is initialized at zero and updated across layers.

To isolate how the test token retrieves and aggregates information from the
context, the controlled nonlinear-attention model in
Section~\ref{sec:model-setup} initially disables context updates by setting
the upper-left block to zero:
\begin{equation}
  \mathsf M(0)
  =
  \begin{pmatrix}
    \bm{0}_{P\times P} & \bm{0}_{P\times 1}\\
    \bm{1}_P^\top & 0
  \end{pmatrix}.
  \label{eq:fixed-context-mask}
\end{equation}

Standard Transformers, however, update context representations across layers.
To examine how these updates affect the nonlinear-attention model, we
parameterize the previously zero context-to-context block by a trainable
scalar $m$. Specifically, we use
\begin{equation}
  \mathsf M(m)=
  \begin{pmatrix}
    m\bm{1}_P\bm{1}_P^\top & \bm{0}_{P\times 1}\\
    \bm{1}_P^\top & 0
  \end{pmatrix},
  \label{eq:trainable-context-mask}
\end{equation}
where the scalar $m$ is shared across all layers and learned jointly with the
other model parameters. 

\begin{figure}[t]
  \centering
  \includegraphics[width=\linewidth]{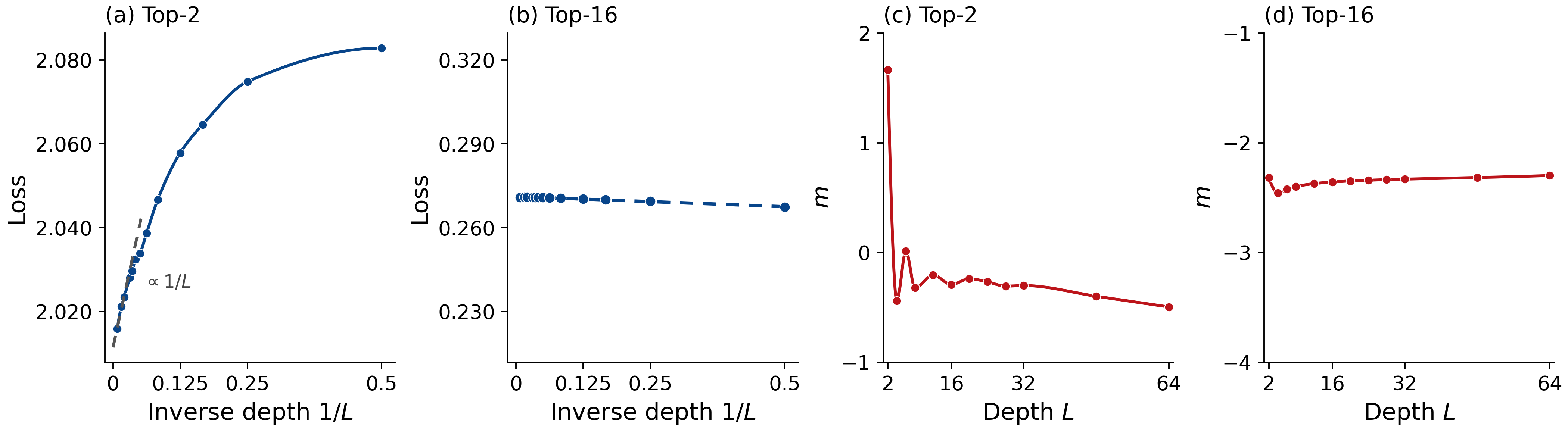}
  \caption{Enabling context updates preserves an inverse-depth scaling signal
  on the top-$2$ task but not on the top-$16$ task. Under isotropic data
  spectra, (a) the top-$2$ loss exhibits an inverse-depth scaling signal,
  whereas (b) the top-$16$ loss quickly saturates.
  (c,d) The learned context-update scalar $m$ varies with
  depth for top-$2$ but stabilizes for top-$16$.}
  \label{fig:context-update-loss-comparison}
\end{figure}

We evaluate this generalized model on the difficult task with isotropic data
spectra, using top-$2$ and top-$16$ targets.
Figure~\ref{fig:context-update-loss-comparison}(a,b) shows qualitatively
different depth dependence. For top-$2$, the loss continues to decrease
substantially with depth and exhibits a signal consistent with inverse-depth
scaling. For top-$16$, by contrast, increasing depth does not improve the
loss. 

The learned context-update scalar $m$ provides an additional diagnostic.
Figure~\ref{fig:context-update-loss-comparison}(c,d) shows that the learned
scalar $m$ varies substantially with depth for top-$2$, whereas for top-$16$
it remains near the stable negative value $-2.3$.

To interpret this task-dependent behavior, we first derive its exact effect in a
two-layer model. Write the label state as
$H_y^\ell=[Z_\ell,h_\ell]$, where $Z_\ell\in\mathbb R^{N\times P}$ contains
the context states and $h_\ell\in\mathbb R^N$ is the test-token state. Let
$y=(y_1,\ldots,y_P)^\top$, so that $Z_0=W_yy^\top$ and $h_0=0$. Without
changing the mechanism of interest we take $W_v=I_N$ for simplicity. Let
$p_\ell\in\mathbb R^{1\times P}$ denote the test-to-context attention row at
layer $\ell$, and let $A_1\in\mathbb R^{P\times P}$ denote the first layer's
context-to-context attention matrix.

Applying \eqref{eq:nonlinear-attention} with the mask in
\eqref{eq:trainable-context-mask} gives
\begin{equation}
 \begin{aligned}
 Z_1&=Z_0+\frac{m}{2}Z_0A_1^\top,\\
 h_1&=\frac12 Z_0p_1^\top,\\
 h_2&=h_1+\frac12 Z_1p_2^\top.
 \end{aligned}
 \label{eq:app-two-layer-state-recursion}
\end{equation}
Substituting $Z_0=W_yy^\top$ gives
\begin{equation}
 h_2
 =W_y\!\left[
 \frac12(p_1+p_2)y+\frac{m}{4}p_2A_1y
 \right].
 \label{eq:app-two-layer-state-expansion}
\end{equation}
Defining the scalar readout gain $g=W_oW_y$, the prediction is
\begin{equation}
 \widehat y_2
 =\frac{g}{2}(p_1+p_2)y+\frac{gm}{4}p_2A_1y.
 \label{eq:app-two-path-context-update}
\end{equation}
The first term reads the original context labels directly. The second reads
them after $A_1$ has mixed information across context tokens.

For arbitrary depth, the corresponding recursion is
\begin{equation}
 \begin{aligned}
 Z_\ell
 &=Z_{\ell-1}\left(I_P+\frac{m}{L}A_\ell^\top\right),\\
 h_\ell
 &=h_{\ell-1}+\frac{1}{L}Z_{\ell-1}p_\ell^\top.
 \end{aligned}
 \label{eq:app-multilayer-context-recursion}
\end{equation}
Consequently,
\begin{equation}
 \widehat y_L
 =\frac{g}{L}\left[
 p_1y+\sum_{\ell=2}^{L}p_\ell
 \left(I_P+\frac{m}{L}A_{\ell-1}\right)\cdots
 \left(I_P+\frac{m}{L}A_1\right)y
 \right].
 \label{eq:app-multihop-context-path}
\end{equation}
Thus, context updates affect prediction through operators
$I_P+(m/L)A_\ell$. Whether and how they improve depth utilization depends on the
spectral structure of $A_\ell$.

\begin{figure}[t]
  \centering
  \includegraphics[width=0.82\linewidth]{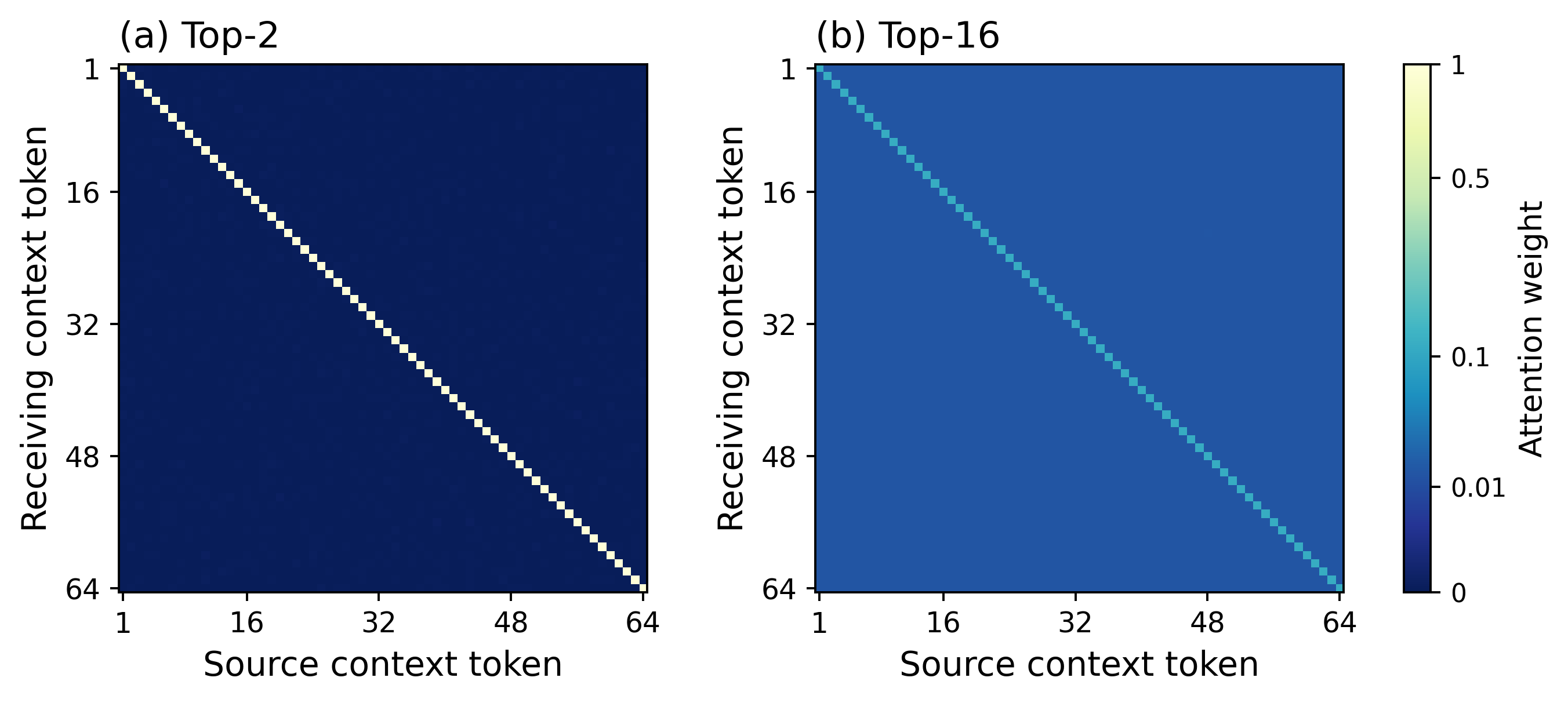}
  \caption{The learned context-attention structure depends strongly on the
  aggregation target. The context-attention matrix $A_1$ is
  (a) sharply concentrated on the diagonal for top-$2$ task, but (b) distributed
  broadly across context tokens for top-$16$ task.}
  \label{fig:context-attention-heatmaps}
\end{figure}

Figure~\ref{fig:context-attention-heatmaps} visualizes the two context attention structures.
For the top-$2$ task, the measured $A_1$ is nearly the identity. For the
top-$16$ task, the measured $A_1$ is broader and flatter. This contrast is
induced by the sharpness of the target distribution. The sharp top-$2$ target
requires the model to preserve the identities of a few selected tokens,
favoring well-separated embeddings and a nearly identity
attention map. The flatter top-$16$ target instead favors stronger coupling
across embeddings and therefore produces a broader attention map.

To quantify this contrast, we model
each measured attention matrix by an identity component $I_P$ and a
uniform mixing component $J_P$. Specifically, for $k\in\{2,16\}$, we fit
$\rho_k$ by least squares in
\begin{equation}
 A_1^{(k)}\approx \rho_k I_P+(1-\rho_k)J_P,
 \qquad
 J_P:=\frac{\bm{1}_P\bm{1}_P^\top}{P}.
 \label{eq:app-context-filter-approximation}
\end{equation}
The identity term preserves each context embedding separately, whereas $J_P$
replaces every embedding by their common average. The fitted results are
\begin{equation}
 \begin{aligned}
  \rho_2&=0.9999, & R_{\mathrm{top}\text{-}2}^2&>0.99999,\\
  \rho_{16}&=0.0989, & R_{\mathrm{top}\text{-}16}^2&=0.99996.
 \end{aligned}
 \label{eq:app-context-filter-fits}
\end{equation}
This model therefore confirms that the
top-$2$ matrix is almost exactly the identity, while the top-$16$ matrix is
dominated by uniform cross-token mixing.

For top-$2$, the near-identity structure gives $p_2A_1\approx p_2$, and
\begin{equation}
 \widehat y_2
 \approx\frac g2p_1y+\frac g2(1+m/2)p_2y.
 \label{eq:app-top2-gain}
\end{equation}
The context update therefore introduces no independent token-weighting
direction: the prediction remains an aggregation of $p_1$ and $p_2$.
Accordingly, its loss remains close to that of the fixed-context model and
still exhibits approximate inverse-depth scaling over the observed depth
range.

For top-$16$, let $u=\bm{1}_P^\top/P$. Since $p_2\bm{1}_P=1$, we have
$p_2J_P=u$ and hence
\begin{equation}
 p_2A_1
 \approx
 \rho_{16}p_2+(1-\rho_{16})u,
 \label{eq:app-top16-diffuse-path}
\end{equation}
Equation~\ref{eq:app-top16-diffuse-path} shows that the context update adds
the uniform token-weighting direction $u$, beyond the direct rows $p_1$ and
$p_2$. Defining $\bar p=(p_1+p_2)/2$, the exact two-layer prediction is
\begin{equation}
 \widehat y_2
 =g\left(\bar p+\frac{m}{4}p_2A_1\right)y.
 \label{eq:app-top16-effective-row}
\end{equation}
A negative $m$ therefore produces a global shift that subtracts the common
attention floor, rather than merely combining more direct attention rows. This
additional degree of freedom allows the model to achieve low loss with few
layers, explaining its early saturation. The same argument extends to
models with more than two layers.

This distinction is also relevant to language modeling. Many language tasks
require focusing on only a few relevant tokens and learning a sparse and peaked
distribution. In such regimes,
context interactions may primarily reweight information that has already been
selected. Depth utilization can therefore remain governed by selective
retrieval and cross-layer aggregation, making the inverse-depth mechanism
relevant even when context representations are not fixed.

Overall, these generalized models exhibit inverse-depth scaling signals that
are qualitatively consistent with the controlled-model analysis over the
accessible depth ranges. Training these models at much greater depths is
substantially more computationally expensive, preventing experiments on the
same scale as those conducted for the controlled model. We therefore interpret
these results as evidence of similar scaling signals, rather than as a
definitive guarantee of asymptotic inverse-depth scaling. Their appearance
under more realistic architectural choices nevertheless suggests that the
mechanism may remain relevant to practical models.

\section{More Experiments and Fitting Details}
\label{app:additional-experiments}

This section supplements the controlled-model results with additional
experiments and provides the fitting procedures used to quantify the observed
depth-scaling laws. We report fits for both the easy and difficult tasks,
analyze context-dependent token selection, test broader targets and data
spectra, and examine the convergence of the shared-error term.

\subsection{Power-Law Fitting Procedures}
\label{app:easy-linear-free-fits}
\label{app:nonlinear-difficult-fits}

We use a common log-space fitting procedure to quantify the depth-scaling
laws in Sections~\ref{sec:easy-linear} and~\ref{sec:difficult-nonlinear}. Each
spectrum is fitted independently using
\begin{equation}
  \mathcal R_L=L_0+cL^{-\alpha},
  \qquad c>0,\quad \alpha>0,
  \label{eq:power-law-fit-general}
\end{equation}
where $L_0$ is the asymptotic loss floor and $\alpha$ is the decay exponent.

For observations at depths $\{L_i\}$, we minimize the squared residuals
between the logarithms of the observed and modeled losses:
\begin{equation}
  (\widehat L_0,\widehat c,\widehat\alpha)
  =\underset{L_0,c,\alpha}{\arg\min}
  \sum_i
  \left[
    \log\mathcal R_{L_i}
    -\log\!\left(L_0+cL_i^{-\alpha}\right)
  \right]^2.
  \label{eq:power-law-logspace-fit-general}
\end{equation}

For the easy task, we set $L_0=0$. As shown in
Appendix~\ref{app:linear-polynomial-theory}, given sufficient depth, width,
dataset size, context length, and training time, the linear-attention model can
solve this task exactly by using a covariance polynomial to approximate the
identity map. For the difficult task, the decomposition in
Appendix~\ref{app:theory} contains an error component shared across layers, so
we estimate $L_0$ jointly with $c$ and $\alpha$.

\subsection{Complementary Token Selection Across Contexts}
\label{app:attention-percentiles}

We further compare how nonlinear-attention models of different depths perform
on the difficult task. We train an $L=1$ model and an
$L=256$ model on the difficult task with $k=2$ under identical conditions. We then sample
$10{,}240$ fresh contexts independently from the data distribution and
evaluate both models on the same set.

For each context, we order the tokens by their similarity to the test token.
Under the parallel interpretation developed in Appendix~\ref{app:theory},
the final output aggregates the contributions of the individual layers. We
therefore summarize their collective token selection by the aggregated attention
distribution and compare it with the target distribution over the ordered
tokens. Formally,
\begin{equation}
  \begin{gathered}
    \bar p_L(c)=\frac{1}{L}\sum_{\ell=1}^{L}p_\ell(c),
    \qquad
    p^*=(1/2,1/2,0,\ldots,0),\\
    E_L(c)=\|\bar p_L(c)-p^*\|_2^2.
  \end{gathered}
  \label{eq:context-selection-loss}
\end{equation}
Lower values indicate more accurate identification and weighting of the two
target tokens. For each model separately, we sort the sampled contexts by
$E_L(c)$ and denote the $q$th percentile by $Q_L(q)$. Thus, small $q$
corresponds to contexts on which a model most closely matches the target
distribution, whereas large $q$ probes its high-error tail.

\begin{figure}[t]
  \centering
  \includegraphics[width=0.95\linewidth]{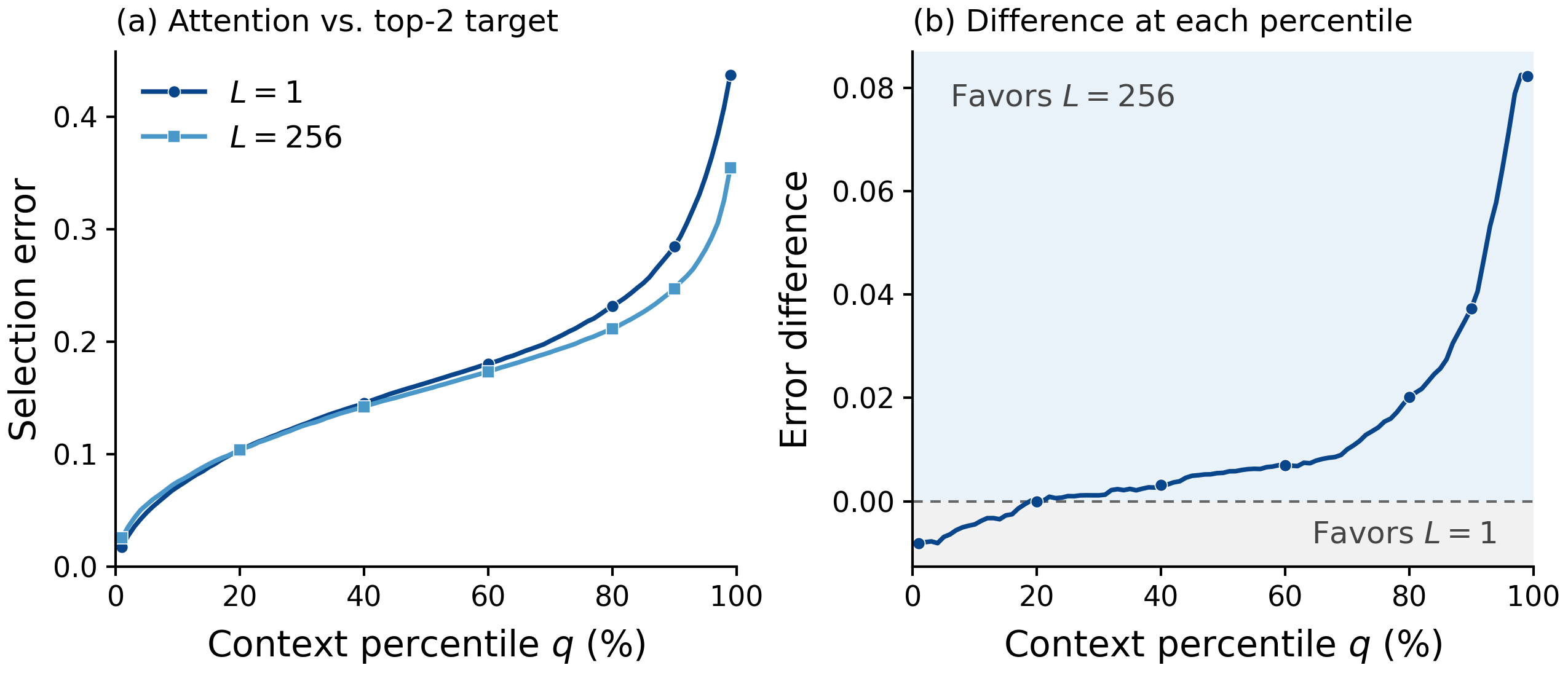}
  \caption{Greater depth makes token selection more reliable across contexts.
  (a) Selection-loss percentiles for $L=1$ and $L=256$.
  (b) Their difference $\Delta Q(q)=Q_1(q)-Q_{256}(q)$; positive values favor
  $L=256$.}
  \label{fig:attention-percentiles}
\end{figure}

Figure~\ref{fig:attention-percentiles}(a) compares the resulting error
distributions. Panel (b) reports
\begin{equation}
  \Delta Q(q)=Q_1(q)-Q_{256}(q),
  \label{eq:context-quantile-improvement}
\end{equation}
Positive $\Delta Q(q)$ therefore indicates that the deeper model has lower
selection loss at percentile $q$. The one-layer model is slightly more accurate on the contexts
in its lowest-error tail, but its error grows much more rapidly across harder
contexts. The $256$-layer model instead markedly reduces the upper tail,
indicating more reliable token selection across the context distribution.

This pattern follows from the fact that each query--key parameterization must
serve the entire context distribution. With only one layer, a single selection
rule may fit some contexts particularly well but cannot identify the correct
top-$2$ tokens uniformly. Greater depth supplies multiple independently
parameterized rules whose errors can be complementary across contexts. Their
aggregation improves coverage: for a context that is difficult for one rule,
other layers may still identify the relevant tokens, thereby reducing the
high-error tail.

The slight advantage of the one-layer model at the lowest percentiles suggests
a possible tradeoff between specialization and coverage. A single selection rule may
become highly accurate on a limited subset of contexts, whereas complementary
rules across layers can cover a broader range. This observation suggests a
testable hypothesis for settings with finitely many or clustered context
types: different layers may specialize in different context classes, and
their aggregation may combine these specialized behaviors. If this hypothesis
holds, it would suggest that current architectures use depth inefficiently,
as not every layer makes an indispensable contribution. Further exploring
designs that exploit such layerwise specialization may help LLMs utilize
depth better.

\subsection{Broader Targets and Data Spectra}
\label{app:additional-spectra}

The main-text experiments with the nonlinear-attention model on the difficult
task use top-$2$ aggregation and vary one spectral parameter while fixing the
other at zero. We now test whether the observed depth scaling depends on either
restriction. First, we retain the top-$2$ target but consider jointly
structured input and task spectra, with both $a$ and $b$ nonzero. Second, we
replace top-$2$ aggregation with top-$16$ aggregation, yielding a flatter
target distribution, and evaluate several spectral settings. For each
configuration, we fit the evaluation loss using the model and log-space
objective defined in Appendix~\ref{app:nonlinear-difficult-fits}.

\begin{figure}[!htbp]
  \centering
  \begin{minipage}{0.49\linewidth}
    \centering
    \includegraphics[width=\linewidth]{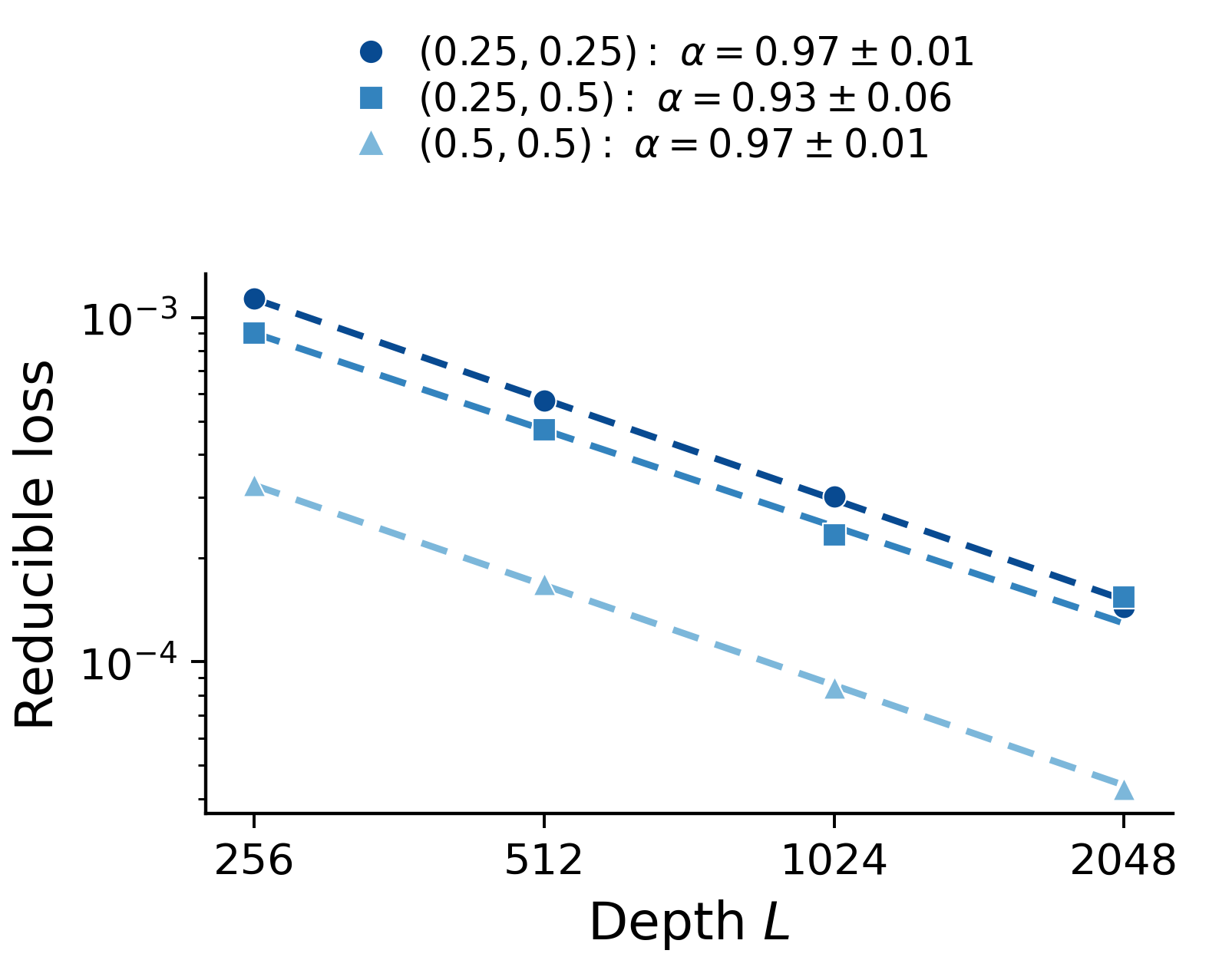}
    \par\small (a) Top-$2$ aggregation
  \end{minipage}\hfill
  \begin{minipage}{0.49\linewidth}
    \centering
    \includegraphics[width=\linewidth]{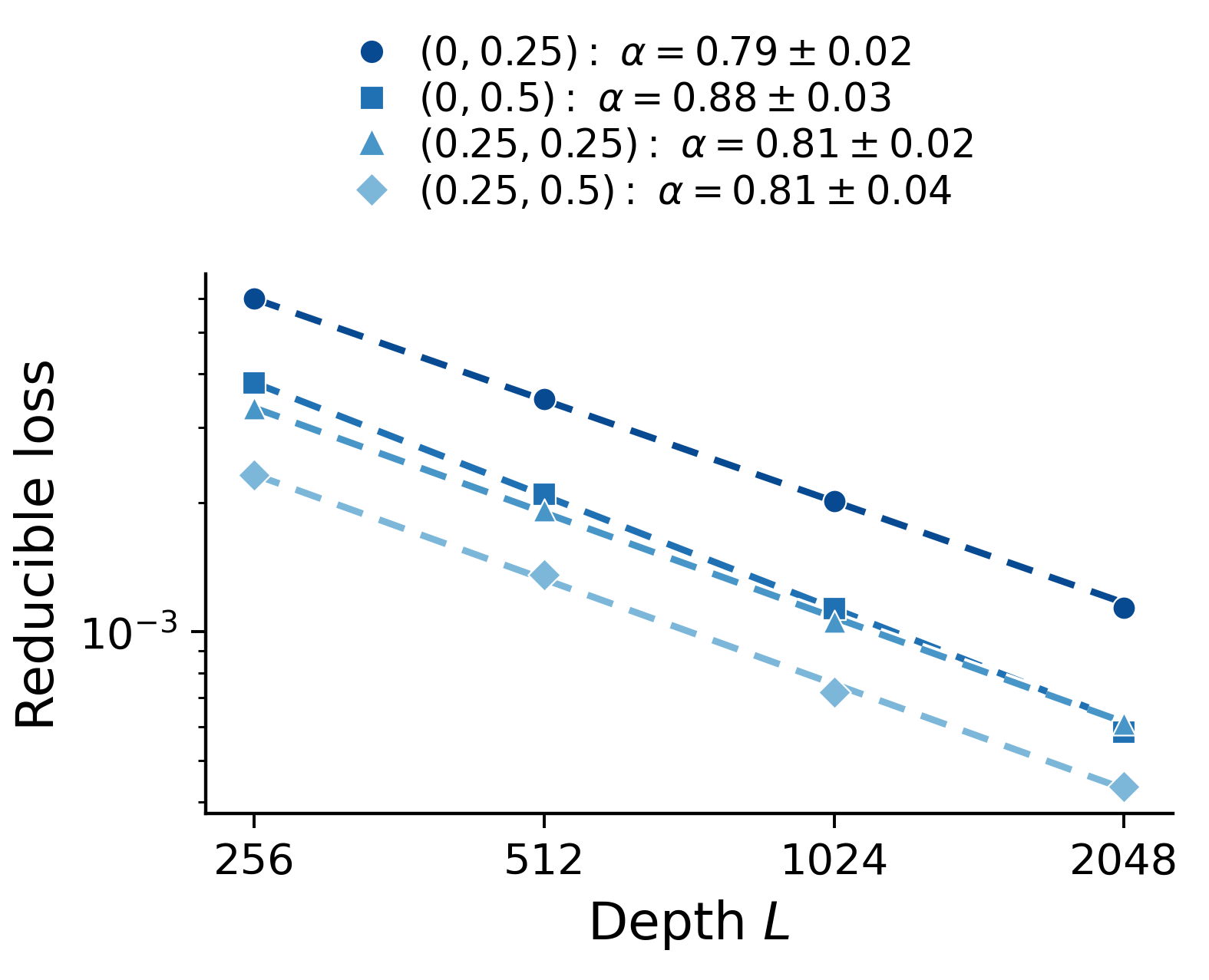}
    \par\small (b) Top-$16$ aggregation
  \end{minipage}
  \vspace{0.5\baselineskip}

  \begin{minipage}{0.49\linewidth}
    \centering
    \includegraphics[width=\linewidth]{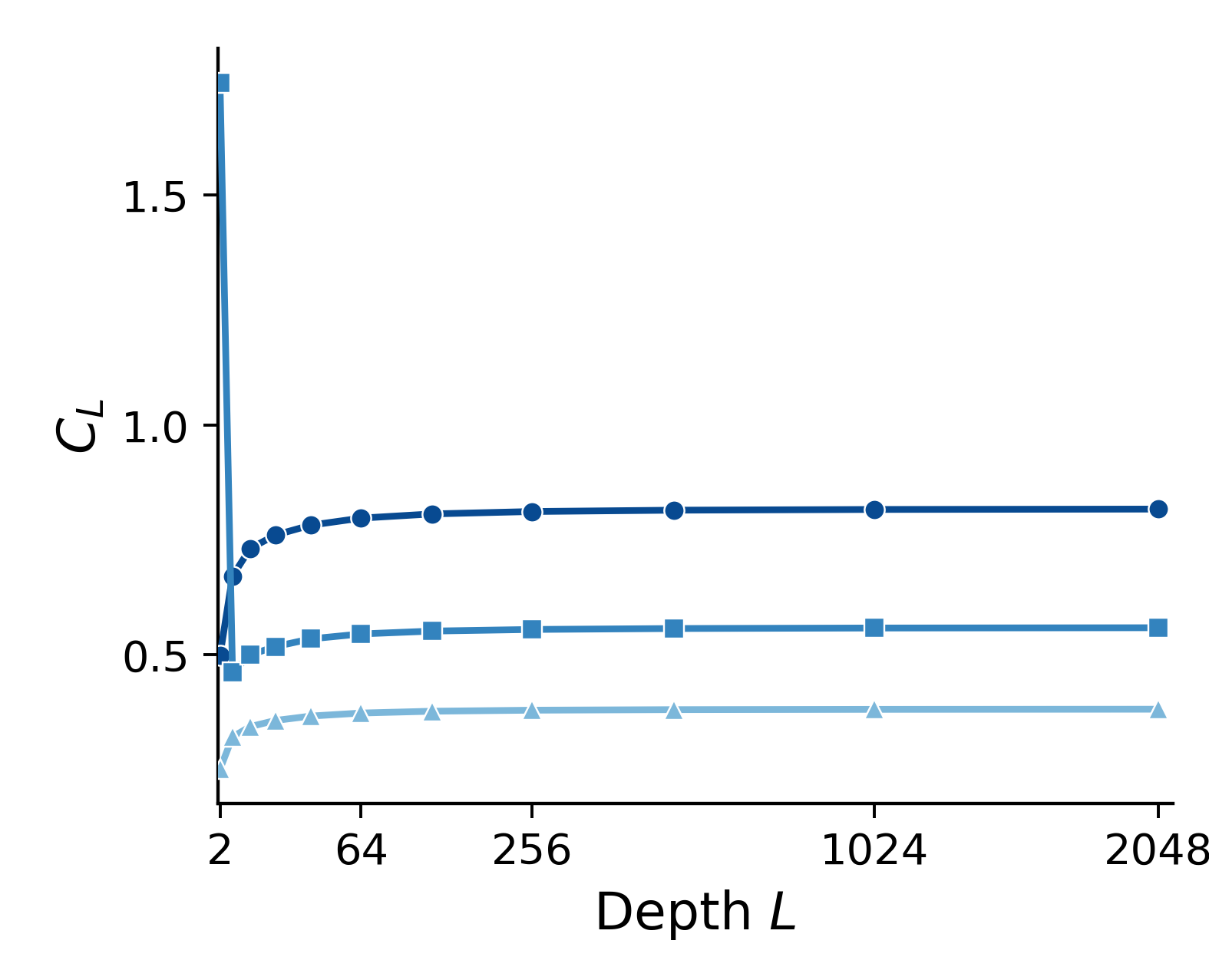}
    \par\small (c) Shared error
  \end{minipage}\hfill
  \begin{minipage}{0.49\linewidth}
    \centering
    \includegraphics[width=\linewidth]{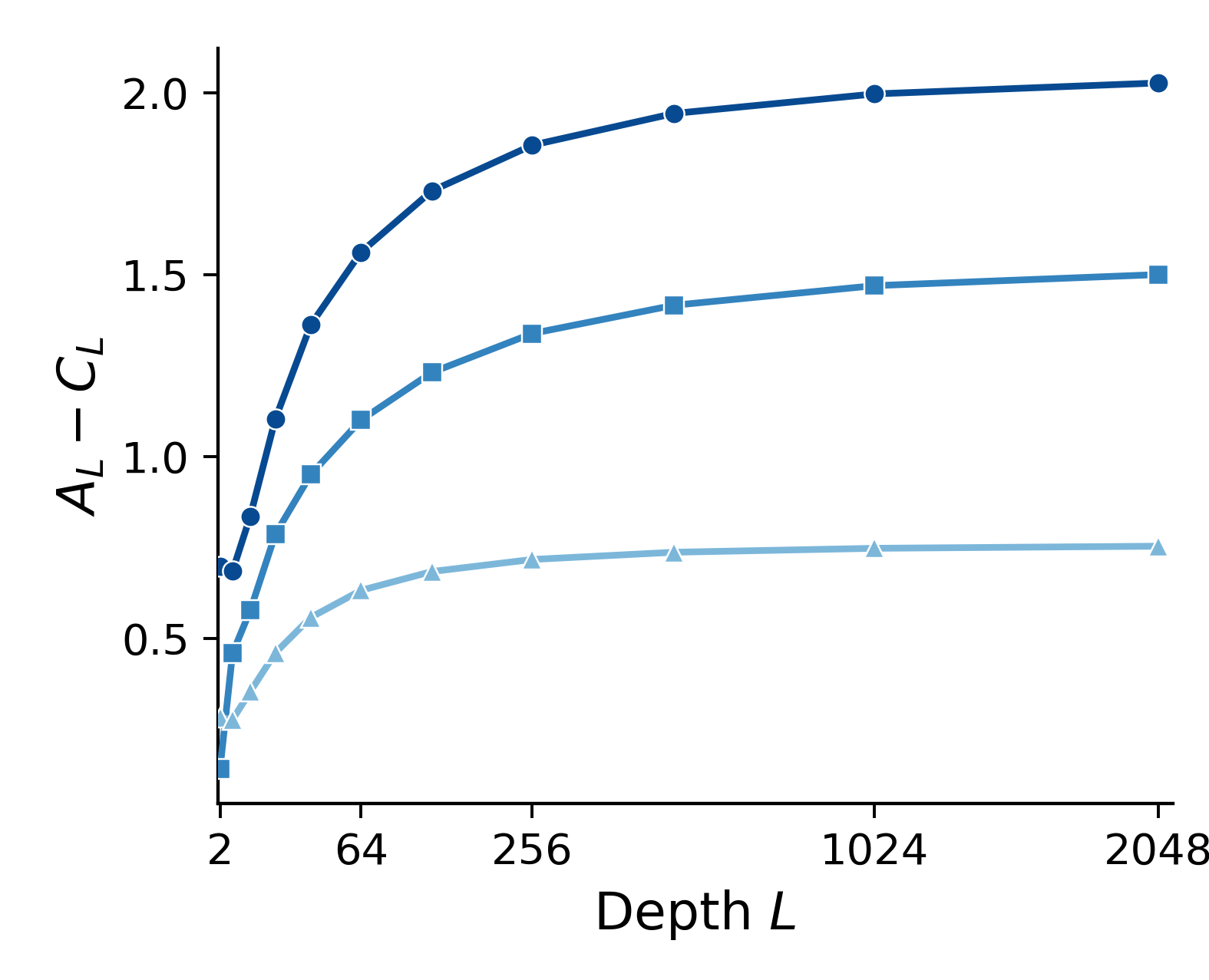}
    \par\small (d) Error disagreement
  \end{minipage}
  \caption{Emergent inverse-depth scaling in the deep regime generalizes across broader
  targets and data spectra. (a) Performance on the top-$2$ task with both spectral
  parameters nonzero. (b) Performance on top-$16$ aggregation across representative
  spectra. (c) $C_L$ and (d) $A_L-C_L$ converge to finite constants with depth for
  the three top-$2$ spectra.}
  \label{fig:appendix-generalization}
\end{figure}

\begin{table}[!htbp]
  \centering
  \caption{Fitted depth-scaling exponents remain close to one across broader
  targets and data spectra.}
  \label{tab:appendix-generalization}
  \vspace{\baselineskip}
  \small
  \setlength{\tabcolsep}{9pt}
  \begin{tabular}{ccrrr}
    \toprule
    Target & $(a,b)$ & $L_0$ & $\alpha\pm\mathrm{SE}$ & $R^2$ \\
    \midrule
    Top-$2$ & $(0.25,0.25)$ & 0.818068 & $0.97\pm0.01$ & 0.9998 \\
            & $(0.25,0.5)$  & 0.559694 & $0.93\pm0.06$ & 0.9984 \\
            & $(0.5,0.5)$   & 0.382188 & $0.97\pm0.01$ & 0.9999 \\
    \midrule
    Top-$16$ & $(0,0.25)$    & 0.173237 & $0.79\pm0.02$ & 0.9999 \\
             & $(0,0.5)$     & 0.114436 & $0.88\pm0.03$ & 0.9998 \\
             & $(0.25,0.25)$ & 0.115248 & $0.81\pm0.02$ & 0.9998 \\
             & $(0.25,0.5)$  & 0.079785 & $0.81\pm0.04$ & 0.9994 \\
    \bottomrule
  \end{tabular}
\end{table}

Figure~\ref{fig:appendix-generalization} shows approximately inverse-depth loss
decay across both extensions, and Table~\ref{tab:appendix-generalization}
quantifies the corresponding exponents. For the jointly nonzero top-$2$
settings, panels
(c,d) further show that both decomposition terms approach spectrum-dependent
finite values, matching the convergence pattern observed in the main text.
With jointly nonzero spectra, the
top-$2$ exponents remain close to one. Note that the
top-$16$ exponents deviate relatively more
from one at the depths considered, suggesting that the top-$16$ task may enter
the asymptotic inverse-depth regime only at greater depths. This contrast also
suggests that inverse-depth scaling from cross-layer aggregation may be more
pronounced for sparse, sharply concentrated target distributions. These results
show that the inverse-depth scaling behavior is not restricted to specific
spectra or target distributions.

\subsection{Convergence of the Shared-Error Term}
\label{app:shared-error-convergence}

The direct loss fits in Section~\ref{sec:difficult-nonlinear} provide empirical
evidence for approximate inverse-depth scaling over the tested depth range.
As a consistency check on the decomposition in Appendix~\ref{app:theory},
we examine how its shared-error term $C_L$ approaches the fitted loss floor.
For each spectrum, we fix $L_0$ to the corresponding estimate in
Table~\ref{tab:main-nonlinear-fits} and model the gap as
\begin{equation}
  L_0-C_L=cL^{-\alpha},
  \qquad c>0,\quad \alpha>0.
  \label{eq:app-shared-error-rate}
\end{equation}
Following the log-space procedure in
Appendix~\ref{app:nonlinear-difficult-fits}, we estimate the two parameters
by least squares:
\begin{equation}
  (\widehat c,\widehat\alpha)
  =\underset{c>0,\,\alpha>0}{\arg\min}
  \sum_i
  \left[
    \log(L_0-C_{L_i})-\log\!\left(cL_i^{-\alpha}\right)
  \right]^2.
  \label{eq:app-shared-error-fit}
\end{equation}

\begin{table}[!htbp]
  \centering
  \caption{The gap $L_0-C_L$ decays approximately inversely with depth
  over the fitted range.}
  \label{tab:shared-error-convergence}
  \vspace{\baselineskip}
  \small
  \begin{tabular}{ccc}
    \toprule
    $(a,b)$ & $\alpha\pm\mathrm{SE}$ & $R^2$ \\
    \midrule
    $(0,0)$    & $0.980\pm0.007$ & 0.99981 \\
    $(0,0.25)$ & $0.983\pm0.038$ & 0.99396 \\
    $(0,0.5)$  & $0.977\pm0.014$ & 0.99923 \\
    $(0,1)$    & $0.982\pm0.011$ & 0.99946 \\
    $(0.25,0)$ & $0.990\pm0.015$ & 0.99904 \\
    $(0.5,0)$  & $0.986\pm0.005$ & 0.99988 \\
    $(1,0)$    & $0.994\pm0.005$ & 0.99992 \\
    \bottomrule
  \end{tabular}
\end{table}

With $L_0$ fixed by the direct loss fits, the high $R^2$ values and small
fitting uncertainties show that $L_0-C_L$ is well described by a power law
over the tested depths. Its fitted decay exponents are close to one, consistent
with the inverse-depth trend observed in the loss.

\end{document}